\documentclass[12pt]{article}
\usepackage{amsthm} 
\usepackage{amssymb}  
\usepackage[english]{babel}
\usepackage[utf8]{inputenc}
\usepackage{fancyhdr}
\usepackage{array}
\usepackage{booktabs}
\usepackage{colortbl}
\usepackage{eurosym}
\usepackage{amsmath}
\usepackage{graphicx}
\usepackage{color}
\usepackage[ruled,lined,linesnumbered,boxed]{algorithm2e}
\usepackage{algorithmic}
\usepackage{bbm,amssymb}
\usepackage{multirow}

\newcommand{\prox}{{\rm prox}}

\newcommand{\Argmin}{\textrm{Argmin}}

\newtheorem{proposition}{Proposition}
\newtheorem{problem}{Problem}

\begin{document}

\title{A deep primal-dual proximal network\\ for image restoration}


\author{Mingyuan Jiu \thanks{School of Information Engineering, Zhengzhou University, Zhengzhou, 450001, China. Email: iemyjiu@zzu.edu.cn} \and
Nelly Pustelnik \thanks{Univ Lyon, Ens de Lyon, Univ Lyon 1, CNRS, Laboratoire de Physique, Lyon, 69342, France. Email: nelly.pustelnik@ens-lyon.fr}
}


\maketitle

\begin{abstract}
Image restoration remains a challenging task in image processing. Numerous methods tackle this problem, which is often  solved by minimizing a nonsmooth penalized co-log-likelihood  function. Although the solution is easily interpretable with theoretic guarantees,  its estimation relies on an optimization process that can take time. Considering the research effort in deep learning for image classification and segmentation, this class of methods offers a serious alternative to perform image restoration but stays challenging to solve inverse problems. In this work, we design a deep network, named DeepPDNet, built from primal-dual proximal iterations  associated with the minimization of a standard penalized co-log-likelihood with an analysis prior, allowing us to take advantage of both worlds. 

We reformulate a specific instance of the Condat-V\~u primal-dual hybrid gradient (PDHG) algorithm as a deep network with fixed layers. Each layer corresponds to one iteration of the primal-dual algorithm. 
The learned parameters are both the PDHG algorithm step-sizes and  the analysis linear operator involved in the penalization (including the regularization parameter). These parameters are allowed to vary from a layer to another one. Two different learning strategies: ``Full learning'' and ``Partial learning'' are proposed, the first one is the most efficient numerically while the second one relies on standard constraints ensuring convergence of the standard PDHG iterations.   Moreover, global and local sparse analysis prior are studied to seek a better feature representation. We apply the proposed methods to image restoration on the MNIST and BSD68 datasets and to single image super-resolution on the BSD100 and SET14 datasets. Extensive results show that the proposed DeepPDNet demonstrates excellent performance on the MNIST dataset compared to other state-of-the-art methods and better or at least comparable performance on the more complex BSD68, BSD100, and SET14 datasets for image restoration and single image super-resolution task. 
\end{abstract}

\section{Introduction} \label{sec:intro}

Inverse problem solving has been studied for many years with applications ranging from astrophysics to medical imaging. Among the numerous methods dedicated to this subject, we can first refer to the pioneering work by Tikhonov~\cite{Tikhonov_A_1963_j-sov-mat-dok_tikhonov_ripp}, studying the stability by introducing a smooth penalization, but also to the fundamental contributions about penalized co-log-likelihood by Geman and Geman~\cite{geman1987:bayesian_resto} in the discrete setting and its continuous counterpart proposed by Mumford and Shah~\cite{mumfordShah1989:Optimal_Approximation}. An important research effort was then related to compressed sensing theory, sparsity, and proximal algorithmic strategies, allowing to solve large size nonsmooth penalized co-log-likelihood objective function~\cite{bauschke_convex_2017,Combettes2011}. Most of the contributions in inverse problems for image analysis between 2000 and 2015 were dedicated to such nonsmooth objective functions leading to major improvements in the reconstruction performance. In parallel, in the domain of image classification and segmentation, outstanding performances have been achieved using deep learning strategies~\cite{LeCun_Y_2015_nature_dee_l}, which offer a promising research direction for solving inverse problems too. However, its counterpart for inverse problems is still an active research area in order to obtain a solution as understandable and stable as the one obtained with the well-studied penalized co-log-likelihood minimization approaches.

In this work, we consider an original image $\overline{\textrm x}\in \mathbb{R}^{N}$ composed with $N$ pixels and its degradation model:
\begin{equation}\label{eq:degrad_model}
{\textrm z} =  A\overline{\textrm x} + \varepsilon
\end{equation}
where $ A\in \mathbb{R}^{M\times N}$ models a linear degradation,  $\varepsilon\sim \mathcal{N}(0,\alpha^2 \mathbb{I})$ models the effect of white Gaussian noise with a standard deviation $\alpha$, and ${\textrm z}\in \mathbb{R}^{M}$ denotes the observed data. The resolution of an inverse problem relies on the estimation of $\widehat{\textrm x}$ from the observations $\textrm z$, and possibly the knowledge of $ A$. Penalized co-log-likelihood approaches rely on the minimization of an objective function being the sum of a data fidelity term (co-log-likelihood) and a penalization (prior), expressed in its most standard formulation as:
\begin{equation}\label{eq:general_prob}
\widehat{\textrm x}_\lambda\in \underset{\textrm x}{\Argmin}\; \frac{1}{2}\Vert  A \textrm x - \textrm z \Vert_{2}^2 + \lambda p( H\textrm x),
\end{equation}
where $\lambda >0$ denotes the regularization parameter allowing a tradeoff between the first term, insuring the solution  to be  \emph{close to} the observations (and mostly designed according to the noise statistics), and the penalization involving a linear operator $ H\in \mathbb{R}^{P\times N}$ and a function $p\colon \mathbb{R}^P\to ]-\infty,+\infty]$. The modelization of the prior as the composition of a function with a linear operator allows us to model most of the standard penalization such as the smooth convex Tikhonov or hyperbolic Total Variation penalization \cite{Charbonnier_P_1997_j-ieee-tip}, the anisotropic or isotropic  Total Variation (TV)  \cite{RudinOsher92PhysD, rudin1994:TV_image_resto_free_local_constraints, chambolle1997:Image_recovery_TV,chambolle2004:Algo_TV, osher2005:Iterative_regularization_TV_restoration,condat2017:discrete_TV} or its generalization referred as non-local TV (NLTV) \cite{peyre2008:nonlocal_reg,li2017:non_local_TV}, wavelet or frame based penalizations in its analysis formulation~\cite{Elad_M_2007_j-ip_ana_vss,Chaari_L_2009_spie_sol_ip}, penalization based on Gaussian mixture models (GMM) such as EPLL \cite{Zoran_D_2017_p-iccv_fro_lmn},   
or BM3D for image denoising~\cite{Danielyan2012}. The reader can refer to \cite{osher2005:Iterative_regularization_TV_restoration,Pustelnik_N_20016_j-w-enc-eee_wav_bid} for detailed overview and comparisons of these penalizations. The choice of the penalization is very dependent on the application due to the expected computational processing time and the structure of the data that can vary a lot from an application to another one.  

A large panel of proximal-based algorithmic strategies have been developed to estimate $\widehat{\textrm x}_\lambda$ \cite{bauschke_convex_2017,Combettes2011,Condat_L_2019}. Among the most standard ones, we can refer to forward-backward~(FB) algorithm \cite{combettes_signal_2005} and related schemes as the iterative soft-thresholding algorithm (ISTA) and its accelerations \cite{Beck_A_2009_j-siam-is_fast_istalip,Chambolle_A_2015_jota_con_ifa}, Douglas-Rachford (DR) algorithm~\cite{Combettes_P_2007_j-ieee-jstsp_dou_rsa}, ADMM or Split bregman iterations for which links have been estabslished with DR~\cite{Setzer_S_2009_ssvm_spl_bad}, and more recently proximal primal-dual schemes (see \cite{Condat_L_2019} for a detailed review), especially PDHG \cite{condat_primal-dual_2013,Vu2013} providing a general algorithm that can be reduced to either FB or DR. In this work we propose to focus on this last mentioned primal-dual scheme~\cite{condat_primal-dual_2013} which  appears flexible enough to solve \eqref{eq:general_prob} without requiring strong assumptions on $A$ neither on $H$ but only $p$ to be convex, lower-semicontinuous and proper. 
The iterations of \cite{condat_primal-dual_2013,Vu2013} in the specific context of \eqref{eq:general_prob} are:
\begin{align}
\begin{cases}
\textrm{x}^{[k+1]} &= \textrm{x}^{[k]} - \tau  A^*( A\textrm{x}^{[k]} - \textrm{z}) - \tau  H^{*} \textrm{y}^{[k]} \\
\textrm{y}^{[k+1]} &= \textrm{prox}_{\sigma\lambda p^{*}} \big(\textrm{y}^{[k]} + \sigma  H (2\textrm{x}^{[k+1]} - \textrm{x}^{[k]}) \big)
\end{cases}
\end{align}
where $\textrm{prox}$ denotes the proximity operator~\cite{bauschke_convex_2017} and $p^{*}$ is the conjugate function of $p$.
Under technical assumptions, especially involving the choice of the step-sizes $\tau$ and $\sigma$, the sequence $(\textrm{x}^{[k]})_{k\in\mathbb{N}}$  is insured to converge to $\widehat{\textrm{x}}_\lambda$. For the challenging question of the selection of the optimal $\lambda$, one can either have recourse to supervised learning by selecting the optimal $\lambda$ using a training database or to unsupervised technique such as the Stein Unbiased Risk Estimator (i.e.~SURE), with successful applications in~\cite{Ammanouil_art_mnras19, Pascal_B_2020_arxiv}. It  relies on
$$
\underset{\lambda}{\textrm{minimize}}\; \mathbb{E}\{\Vert \phi (\widehat{\textrm{x}}_\lambda - \overline{\textrm{x}})\Vert^2\},
$$
and on a reformulation of this expression which does not involve the knowledge of ground-truth $\overline{\textrm{x}}$, leading either to the estimation risk when $\phi = \textrm{I}$ (can be handled  in the context of inverse problems with full rank $A$) or a prediction risk  that is $\phi =  A$ (for more general linear degradation)~\cite{ramani2008monte, eldar2008generalized, Pesquet_J_2009_j-ieee-tsp_sure_ads, deledalle2014stein}.

A recent alternative to variational formulation and nonsmooth optimization relies on supervised learning based on neural networks (see~\cite{Lucas_A_2018_j-ieee-spm,Ravishankar_S_2019_arxiv} for review papers). The contributions dedicated to this subject are wide but the common point is to learn a prediction function $f_\Theta$ with a set of parameters $\Theta$ from a  training data set $\mathcal{S} = \{ (\overline{\textrm{x}}_s, \textrm{z}_s) | s=1, \ldots, I\}$ by minimizing an  emprirical loss function of the form: 
$$
{E}(\Theta):=\frac{1}{I}\sum_{s=1}^I\Vert f_\Theta(\ell_z(\textrm{z}_s)) - \ell_x(\overline{\textrm{x}}_s) \Vert^2.
$$
The simplest approach consists of considering  $f_\Theta$ as a CNN \cite{ Mao_X_2016_nips_usi_dnn, Ulyanov_D_2020_ijcv_dee_cnn,Xu_L_2014_p-nips_dee_cnn}, leading  to $\ell_z(\textrm{z}) = \textrm{z}$ and $\ell_x(\textrm{x}) = \textrm{x}$. Improved performance can be achieved when the learning is performed on wavelet or frame coefficients \cite{Kang_E_2017_mp_deep_cnn, Ye_J_2018_j-siims_dee_cfg, Bubba_T_2019_j-ip_lea_ihd} (leading to $\ell_z$ and  $\ell_x$ associated with the frame transform),  and/or on backprojected data i.e. $\ell_z(\textrm{z}) = A^\dag \textrm{z}$ (or also alternative relying on $\ell_z(\textrm{z}) = A^* \textrm{z}$) \cite{Jin_K_2017_j-ieee-tip_deep_cnn}.  More recently, the design of $f_\Theta$ relies on the knowledge built for many years in inverse problems, for instance by truncated a Neumann series \cite{Gilton_D_2019_j-ieee-tci_neu_nli}, or by unfolding iterative methods such as gradient descent scheme in \cite{RenZuoZhangPAMI2019}, ISTA iterations as proposed in the pioneer work by Gregor and LeCun~\cite{Lecun_2010_icml_lfasc},  proximal interior point iterations as in \cite{Bertocchi_C_2020_j-ip_deep_upp}, and more recently by unfolding a proximal primal-dual optimization method \cite{Adler_J_2018_j-ieee-tmi_lea_pdr} where proximal operators have been replaced with CNN (see also \cite{Zhang_K_2017_p-ccvpr_lea_dcn} for similar ideas). The last class of approaches, especially~\cite{Bertocchi_C_2020_j-ip_deep_upp}, offers a framework particularly suited for stability analysis~\cite{Combettes_P_2020_svva_dee_upi}. 

Our contributions are first to reformulate a specific instance of the Condat-V\~u primal-dual hybrid gradient (PDHG) algorithm applied to solve \eqref{eq:general_prob} as a deep network with fixed layers. Each layer corresponds to one iteration of the primal-dual algorithm (Section~\ref{sec:primaldual}). Based on this relation, a second contribution consists in reformulating primal-dual algorithm into a deep network framework aiming to learn both the algorithmic parameters $\sigma$, $\tau$ and also the regularization parameter $\lambda$ and the linear operator $H$ (as a unique entity $\lambda H$) for a fixed number of layers $K$, leading to the proposed DeepPDNet (Section~\ref{sec:dpdn}). Then, we design a backpropagation procedure based on an explicit differentiation with respect to the parameters of interest.  Global and local sparse analysis prior are  studied to seek better feature representation (Section~\ref{s:gls}). Finally, the proposed network is evaluated on image restoration and single image super-resolution problems with different levels of complexity in terms of noise, blur, and database (Section~\ref{sec:experiment}). \\

\section{Neural networks versus Primal-dual proximal scheme} \label{sec:primaldual}

\subsection{Condat-V\~u algorithm}
The design of our neural network  relies on a reformulation of \eqref{eq:general_prob}, which is summarized in Problem~\ref{prob:1}.
\begin{problem}\label{prob:1} Let $A\in \mathbb{R}^{M\times N}$,   $\mathrm{z}\in \mathbb{R}^M$, $L\in \mathbb{R}^{P\times N}$, and $g$ a convex, l.s.c, and proper function from $\mathbb{R}^P$ to $]-\infty,+\infty]$ such that
	\begin{equation}\label{eq:basic_prob}
	\widehat{\mathrm x}_\lambda\in \underset{\mathrm x \in \mathbb{R}^N}{\mathrm{Argmin}} \;\frac{1}{2}\Vert  A \mathrm x - \mathrm z \Vert_{2}^2+ g(L\mathrm x).
	\end{equation}
\end{problem}
A particular instance of this problem is provided in~\eqref{eq:general_prob} where  $g\circ L = \lambda  p( H \cdot)$. In the specific case where $p = \Vert \cdot \Vert_1$, then $L = \lambda H$, which allows us to combine the estimation of $\lambda$ and $H$.

In order to  solve the nonsmooth objective function~\eqref{eq:basic_prob} in a general setting without specific assumptions on $L$ and $A$ (e.g.  a tight frame or a matrix associated with a filtering operation with periodic boundary effects), the most flexible and intelligible algorithm in the literature is certainly the Condat-V\~u algorithm~\cite{condat_primal-dual_2013}, whose iterations specified to  the resolution of Problem~\ref{prob:1} are summarized in Algorithm~\ref{algo:primaldual}.

\begin{algorithm}[h] 
	\textbf{Set:} $\tau>0$, $\sigma>0$, such that $\frac{1}{\tau} - \sigma\Vert L \Vert^2 > \frac{\Vert A \Vert^2}{2}$ 
	
	\textbf{Initialization:} $(\textrm{x}^{[1]}, \textrm{y}^{[1]})\in \mathcal{H}\times \mathcal{G}$ 
	
	\For{$k=1, \ldots K$}{ 
		
		$\textrm{x}^{[k+1]} = \textrm{x}^{[k]} - \tau  A^*(A\textrm{x}^{[k]} - \textrm{z}) - \tau L^{*} \textrm{y}^{[k]}$ 
		
		$\textrm{y}^{[k+1]} = \textrm{prox}_{\sigma g^{*}} \big(\textrm{y}^{[k]} + \sigma L (2\textrm{x}^{[k+1]} - \textrm{x}^{[k]}) \big)$
	}
	\caption{Primal-dual splitting algorithm for solving Problem~\ref{prob:1}.} \label{algo:primaldual}
\end{algorithm}

The core ingredient of proximal algorithms is the proximal operator defined as:
$$(\forall \textrm{x}\in \mathcal{H}) \quad \textrm{prox}_g(\textrm{x}) = \arg \min_{\textrm{y}\in \mathcal{H}}  {\frac{1}{2}}\Vert \textrm{y} - \textrm{x} \Vert_2^2 + g(\textrm{y}).$$
Several properties of the proximity operator have been established in the literature (see \cite{bauschke_convex_2017,Combettes2011} and reference therein).  Among them, the Moreau identity allows us to provide a relation between a function $g$ and its conjugate $g^*$: $\textrm{prox}_{\sigma g^{*}}(\textrm{x}) = \textrm{x} - \sigma \textrm{prox}_{g/\sigma}(\textrm{x}/\sigma)$ for some $\sigma>0$. 

The convergence  of the sequence $(\textrm{x}^{[k+1]})_{k\in\mathbb{N}}$, generated by Algorithm~\ref{algo:primaldual}, to a minimizer of~\eqref{eq:basic_prob} is ensured under a technical assumption involving the step-size parameters $\tau$, $\sigma$, but also $\Vert A \Vert^2$ and $\Vert L \Vert^2$: 
\begin{equation}\label{eq:condpd}
\frac{1}{\tau} - \sigma\Vert L \Vert^2 > \frac{\Vert A \Vert^2}{2}.
\end{equation} 

\subsection{Deep Primal Dual Network} \label{sec:deepnet}

Deep networks are composed of a stack of layers. Each layer takes the output of the previous layer as input and obtains the feature maps after linear transforms (e.g. convolution) and nonlinear activation functions.  Formally,  a network with $K$ layers can be  written as
\begin{equation}
\textrm{u}^{[K]} = \eta^{[K]} \big(D^{[K]}\ldots \eta^{[1]} (D^{[1]}\textrm{u}^{[1]} + b^{[1]})\ldots + b^{[K]}\big) \label{equa:reform}
\end{equation}
where, for every $k\in \{1,\ldots,K\}$,   $D^{[k]}$ is the weight matrix, $b^{[k]}$ is the bias, and $\eta^{[k]}$ is the nonlinear activation function. In the classical deep learning framework (e.g.~CNNs), $D^{[k]}$ is regarded as the convolution operation with a collection of small filters and each filter is associated with a bias $b^{[k]}$, $\eta^{[k]}$ is a nonlinear activation function such as $\textrm{tanh}$, $\textrm{sigmoid}$ or ReLU function, followed by a pooling layer (e.g~max pooling, average pooling, etc.) that is acted to increase local receptive fields and to decrease the parameter number of the network as well.

Proposition~\ref{prop:main} gives the iterations of Condat-V\~u primal-dual algorithm into the deep neural network  formalism \eqref{equa:reform}.

\begin{proposition}\label{prop:main}
	Algorithm~\ref{algo:primaldual} fits the network \eqref{equa:reform} when considering, for every $k\in \{1,\ldots,K\}$, $D^{[k]} \in \mathbb{R}^{(N+P)\times (N+P)}$, $b^{[k]}  \in \mathbb{R}^{N+P}$ and $\eta^{[k]} \colon \mathbb{R}^{N+P} \to  \mathbb{R}^{N+P} $ such that
	{\small{\begin{equation}\begin{cases}
			D^{[k]} = \begin{pmatrix}
			\mathrm{Id} - \tau A^*A & - \tau L^*\\
			\sigma L(\mathrm{Id} - 2\tau A^*A) &\mathrm{Id} - 2\tau \sigma L L^* \\
			\end{pmatrix}\\
			b^{[k]} =  \begin{pmatrix}
			\tau A^*\mathrm{z}\\
			2\tau  \sigma LA^*\mathrm{z}\\
			\end{pmatrix}\\
			\eta^{[k]} =  \begin{pmatrix}
			\mathrm{Id}\\
			\mathrm{prox}_{\sigma g^*}\\
			\end{pmatrix}\\
			\end{cases} \label{equa:paramdef}\end{equation}}}
	where $\mathrm{Id}$ denotes the identity matrix.
\end{proposition}

\begin{proof}
	The result is straightforward setting $$\textrm{u}^{[k]} = \begin{pmatrix}\textrm{x}^{[k]}\\ \textrm{y}^{[k]} \end{pmatrix} \in \mathbb{R}^{N+P}$$ and from the rewriting of primal-dual updates  
	as
	$$\begin{cases}
	\textrm{x}^{[k+1]} = (\textrm{Id} - \tau A^*A)\textrm{x}^{[k]} - \tau L^*  \textrm{y}^{[k]}+\tau A^*\textrm{z} \\
	\textrm{y}^{[k+1]} = \textrm{prox}_{\sigma g^*}\big( \sigma L(\textrm{Id} - 2\tau A^*A)\textrm{x}^{[k]} +\\ \qquad \qquad\qquad \qquad(\textrm{Id} - 2\tau \sigma L L^*)\textrm{y}^{[k]}+ 2\tau  \sigma LA^*\textrm{z}\big).
	\end{cases}$$
\end{proof}

\begin{figure*}[tbp]
	\centering
	\includegraphics[width=\linewidth]{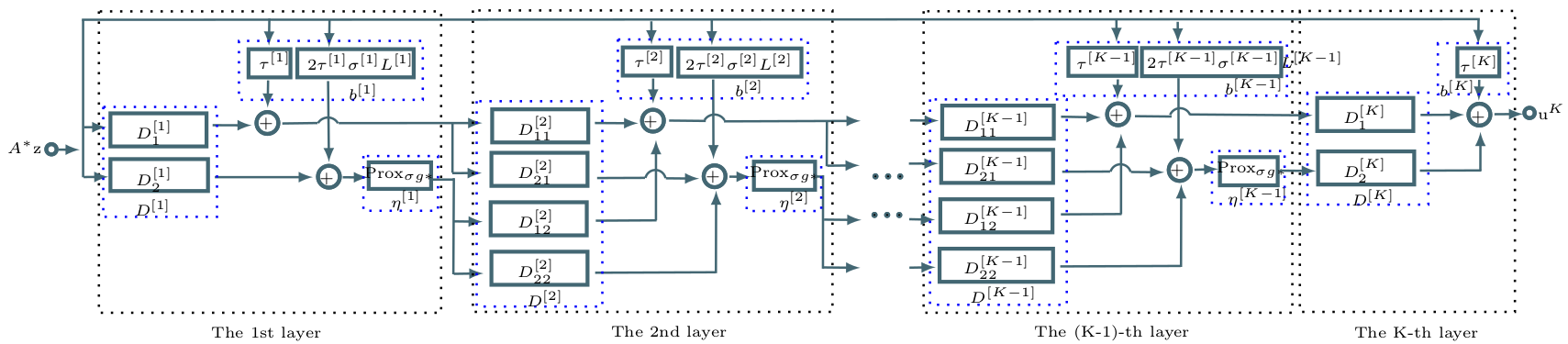}
	\caption{Illustration of the prediction function involved in the proposed DeepPDNet with $K$ layers. In the figure, according to Eq.~\eqref{eq:newstructure},  $D^{[k]}_1=\textrm{Id} - \tau^{[k]} A^* A, k= 1, K$,  $D^{[1]}_2=\sigma^{[1]} L^{[1]}(\textrm{Id} - 2\tau^{[1]} A^* A)$, $D^{[K]}_2=-\tau^{[K]} (L^{[K]})^*$, and for every $ k=\{2,\ldots K-1\}$, $D^{[k]}_{11}=\textrm{Id} - \tau^{[k]} A^* A, D^{[k]}_{12} = -\tau^{[K]} (L^{[K]})^*, D^{[k]}_{21}=\sigma^{[k]} L^{[k]}(\textrm{Id} - 2\tau^{[k]} A^* A), D^{[k]}_{22}=\textrm{Id} - 2\tau^{[k]} \sigma^{[k]} L^{[k]} (L^{[k]})^*$. \label{fig:framework}}
\end{figure*}

Proposition~\ref{prop:main}  presents how to unfold the specific instance of Condat-V\~u primal-dual algorithm described in Algorithm~\ref{algo:primaldual} into a network with multiple layers. This network performs unsupervised image restoration when the number of layers $K\to +\infty$ and it converges to the solution of Eq.~\eqref{eq:basic_prob}. Practically, $K$ needs to be set sufficiently large and its value is either set manually (e.g. $K> 10^5$) or based on a stopping criterion (e.g. based on residual of the iterates or on the duality gap).  

When being reformulated into a supervised learning framework, an extremely large number of $K$ becomes impractical and a deep network with a medium depth (e.g.~ten or hundred layers), that could be interpreted as early stopping, would not provide such a good estimate. So we make several modifications in the network presented in Proposition~\ref{prop:main} suited to a deep learning formalism.

\subsection{Choice of  $L$}
Given Problem~\ref{prob:1}, the  linear operator $L$ is regarded as a prior knowledge (e.g. horizontal/vertical  finite-difference operator, frame transform, \ldots ). Although the merit of the primal-dual splitting algorithm is capable of converging to a fix-point solution, two issues are worthy to be taken into account: on one hand, $L$ is manually set according to prior knowledge, and it is not suitable for different types of datasets, and this may result to poor performance; on the other hand, for each new data, the inference of~\eqref{eq:basic_prob} will take many iterations to ensure the solution to be achieved, which may become impractical for large data. 

In the next section, we deal with these two issues by reformulating the primal-dual algorithm into a deep network framework aiming to learn both the algorithmic parameters $\sigma$, $\tau$, and also the linear operator $L$ for a fixed number of layers $K$.

\section{DeepPDNet : Deep primal dual network}
\label{sec:dpdn}

\subsection{Proposed supervised DeepPDNet}
\label{sec:DeepPDNet}

In a context of restoration where $A$ is known, the degrees of freedom in the proposed network (cf. Proposition~\ref{prop:main}) may only come from $\sigma$, $\tau$, $L$, and $g$. 

In this work, the penalization $g$ is fixed but we let freedom on $L$, including on its norm, leading to implicit freedom on the regularization parameter trade-off. The step-size parameters of the algorithm $\sigma$ and $\tau$ are also learned.  Additionally,  it is assumed that the parameters may be different from a layer to another one, leading to $\Theta = \{\sigma^{[k]}, \tau^{[k]}, L^{[k]}\}_{1 \leq k \leq K}$. 

Given the training set $\mathcal{S} = \{ (\overline{\mathrm{x}}_s, \mathrm{z}_s) | s=1, \ldots, I\}$ where $\overline{\mathrm{x}}_s$ is the original image and $\mathrm{z}_s$ is its degraded counterpart following degradation model~\eqref{eq:degrad_model}. The proposed supervised learning strategy, named DeepPDNet, allows us to learn the parameters $\widehat{\Theta} = \{\widehat{\sigma}^{[k]}, \widehat{\tau}^{[k]}, \widehat{L}^{[k]}\}_{1\leq k \leq K}$ being a solution of:
\begin{equation} \label{eq:newlossk}
\underset{\Theta}{\textrm{minimize}} \;E(\Theta):= \frac{1}{I}\sum_{s=1}^I \Vert \overline{\mathrm{x}}_s - f_\Theta(\mathrm{z}_s) \Vert_2^2
\end{equation}
where the prediction function is
$$
f_\Theta(\mathrm{z}_s) = \eta^{[K]} \big(D^{[K]}(\ldots \eta^{[1]} (D^{[1]}\mathrm{u}_s^{[1]} + b^{[1]})\ldots + b^{[K]}\big) 
$$
with
{\small{\begin{equation} 
		\begin{cases}
		\mathrm{u}_s^{[1]}=A^*\mathrm{z}_s\\
		D^{[1]} = 
		\begin{pmatrix}
		\mathrm{Id} - \tau^{[1]} A^* A\\
		\sigma^{[1]} L^{[1]}(\mathrm{Id} - 2\tau^{[1]} A^* A)\\
		\end{pmatrix}\\
		D^{[k]}\!\!  = \!\! \begin{pmatrix}
		\mathrm{Id} - \tau^{[k]} A^*A & - \tau^{[k]} (L^{[k]})^*\\
		\sigma^{[k]} L^{[k]}(\mathrm{Id} - 2\tau^{[k]} A^*A) &\mathrm{Id} - 2\tau^{[k]} \sigma^{[k]} L^{[k]} (L^{[k]})^* \\
		\end{pmatrix}\\
		b^{[k]} =  \begin{pmatrix}
		\tau^{[k]} A^*\mathrm{z}\\
		2\tau^{[k]}  \sigma^{[k]} L^{[k]}A^*{\mathrm{z}_s}\\
		\end{pmatrix}\\
		\eta^{[k]} =  \begin{pmatrix}
		\mathrm{Id}\\
		\prox_{\sigma^{[k]} g^*}\\
		\end{pmatrix}\\
		D^{[K]} = 
		\begin{pmatrix}
		\mathrm{Id} - \tau^{[K]} A^* A & \tau^{[K]} (L^{[K]})^*\\
		\end{pmatrix}\\ 
		b^{[K]} = \tau^{[K]}A^*{\mathrm{z}_s} \\
		\eta^{[K]} =  \mathrm{Id}.
		\end{cases} \label{eq:newstructure}
		\end{equation}}}

The learning  function $f_\Theta$ is formed with layers which are a direct extension of the layers described in Proposition~\ref{prop:main} where $\sigma$, $\tau$, and $L$ are replaced with parameters depending of the layer $k$ i.e. $\sigma^{[k]}$, $\tau^{[k]}$, $L^{[k]}$. A specific attention needs to be paid due to the primal and dual  input and output involved in the Condat-V\~u scheme. Indeed, the primal-dual algorithm outputs both a primal and a dual solution, while for the output of the network, we do not know the target solution for the dual variable, then we cannot handle the dual solution in the objective function. Thus, if a standard choice for the primal and dual variables setting is $\textrm{x}^{[1]} =A^{*}\textrm{z}_s$ and $\textrm{y}^{[1]} = LA^{*}\textrm{z}_s$, in order to fit \eqref{eq:newlossk}, the initialization of the dual variable is set to $\textrm{y}^{[1]} = 0$. The last layer also needs to be modified in order to only extract the primal variable.

\subsection{Backpropagation procedure}
\label{ss:back}
\noindent The most standard strategy to estimate the parameters in a neural network relies on a stochastic gradient descent algorithm where the objective function is $E(\Theta)$ defined in \eqref{eq:newlossk}. The increments of the parameters are computed from the data (mini-batch strategy is adopted in practice) after forwarding the data through the network. The increments at iteration $\ell+1$ consist of 
$$
\Theta_{[\ell+1]} = \Theta_{[\ell]} - \gamma\nabla E \left(\Theta_{[\ell]}\right),
$$
relying on the updates of each parameter, for every $k \in \{1,\ldots,K\} $,
\begin{equation}
\begin{cases}
\tau^{[k]}_{[\ell+1]} &= \tau^{[k]}_{[\ell]} - \gamma \frac{\partial E}{\partial \tau^{[k]}_{[\ell]}}  \\
\sigma^{[k]}_{[\ell+1]} &= \sigma^{[k]}_{[\ell]} - \gamma \frac{\partial E}{\partial \sigma^{[k]}_{[\ell]}}\\
L^{[k]}_{[\ell+1]} &= L^{[k]}_{[\ell]} - \gamma \frac{\partial E}{\partial L^{[k]}_{[\ell]}} 
\end{cases}  \label{eq:update}
\end{equation}
where $\gamma>0$ is learning step. In order to obtain the gradients in the different layers, we employ a backpropagation procedure, i.e. the errors are backpropagated from last layer to the first layer.

To make clear the presentation of the estimation of $\Theta$, we first rewrite the network feedforward procedure from the layer $k$ to layer $k+1$ as follows:
\begin{equation} \label{eq:forward1}
\textrm{c}^{[k]}_s = D^{[k]} \textrm{u}^{[k]}_s + b^{[k]}
\end{equation}
\begin{equation} \label{eq:forward2}
\textrm{u}^{[k+1]}_s = \eta^{[k]} (\textrm{c}^{[k]}_s).
\end{equation}
For input data $\textrm{z}_s$, the forward procedure to obtain $\textrm{u}^{[K]}_s = f_\Theta(\textrm{z}_s) $ of the network is described in Algorithm~\ref{algo:forwardk}.

\begin{algorithm}[h]
	\small
	\caption{Forward procedure} \label{algo:forwardk}
	\BlankLine
	\KwIn{Set $D^{[k]}$, $b^{[k]}$, $\eta^{[k]}$, $k=\{1,\ldots K\}$ according to \eqref{eq:newstructure}.}
	\KwData{Set $\textrm{u}^{[1]}_s = A^*\textrm{z}_s, s=\{1, \ldots, I\}$}
	\For{$k = 1, \ldots, K$}{
		Perform linear transformation by Eq.~\eqref{eq:forward1}: $\textrm{c}^{[k]}_s = D^{[k]} \textrm{u}^{[k]}_s + b^{[k]}$; \\
		Perform nonlinear activation function by Eq.~\eqref{eq:forward2}: $\textrm{u}^{[k+1]}_s = \eta^{[k]} (\textrm{c}^{[k]}_s)$; \\
	}
\end{algorithm}

Since $\tau^{[k]}$, $\sigma^{[k]}$, and $L^{[k]}$ are jointly involved in $D^{[k]}$, $b^{[k]}$ and $\eta^{[k]}$, so their gradients at iteration $\ell$ read:

\begin{align}
\frac{\partial E}{\partial \tau^{[k]}_{[\ell]}} & = \frac{\partial E}{\partial b^{[k]}_{[\ell]}} \frac{\partial b^{[k]}_{[\ell]}}{\partial \tau^{[k]}_{[\ell]}} + \frac{\partial E}{\partial D^{[k]}_{[\ell]}} \frac{\partial D^{[k]}_{[\ell]}}{\partial \tau^{[k]}_{[\ell]}} \label{eq:taugrad} \\
\frac{\partial E}{\partial \sigma^{[k]}_{[\ell]}} & = \frac{\partial E}{\partial \textrm{u}^{[k+1]}_{s}} \frac{\partial \textrm{u}^{[k+1]}_{s}}{\partial \sigma^{[k]}_{[\ell]}} + \frac{\partial E}{\partial b^{[k]}_{[\ell]}} \frac{\partial b^{[k]}_{[\ell]}}{\partial \sigma^{[k]}_{[\ell]}} + \frac{\partial E}{\partial D^{[k]}_{[\ell]}} \frac{\partial D^{[k]}_{[\ell]}}{\partial \sigma^{[k]}_{[\ell]}} \label{eq:sigmagra}  \\
\frac{\partial E}{\partial L^{[k]}_{[\ell]}} & = \frac{\partial E}{\partial b^{[k]}_{[\ell]}} \frac{\partial b^{[k]}_{[\ell]}}{\partial L^{[k]}_{[\ell]}} + \frac{\partial E}{\partial D^{[k]}_{[\ell]}} \frac{\partial D^{[k]}_{[\ell]}}{\partial L^{[k]}_{[\ell]}}  \label{eq:Lgrad}
\end{align}
\noindent where the gradients of $E$ w.r.t. $D^{[k]}_{[\ell]}$ and $b^{[k]}_{[\ell]}$ are then computed as:
\begin{align}
\frac{\partial E}{\partial D^{[k]}_{[\ell]}} & =  \frac{\partial E}{\partial \textrm{c}^{[k]}_s}  \frac{\partial \textrm{c}^{[k]}_s}{\partial D^{[k]}_{[\ell]}} =\frac{\partial E}{\partial \textrm{c}^{[k]}_s}(\textrm{u}^{[k]}_s)^{\top} \label{eq:gradientDk}\\
\frac{\partial E}{\partial b^{[k]}_{[\ell]}} & = \frac{\partial E}{\partial \textrm{c}^{[k]}_s}. \label{eq:gradientBk}
\end{align}
and where the errors for the variable $\textrm{c}^{[k]}_s$ and $\textrm{u}^{[k]}_s$ at layer $[k]$ are obtained according to Eq.~\eqref{eq:forward2} and Eq.~\eqref{eq:forward1} by chain rule:
\begin{align}
\frac{\partial E}{\partial \textrm{c}^{[k]}_s} & = \frac{\partial E}{\partial \textrm{u}^{[k+1]}_s} \frac{\partial \textrm{u}^{[k+1]}_s}{\partial \textrm{c}^{[k]}_s} \label{eq:derivac} \\
\frac{\partial E}{\partial \textrm{u}^{[k]}_s} & = D^{[k]}_{[\ell]} \frac{\partial E}{\partial \textrm{c}^{[k]}_s},  \label{eq:derivauk}
\end{align}
relying on the error of loss $E$ w.r.t. $\textrm{u}^{[k+1]}_s$ at the layer $k+1$ (denoted as $\frac{\partial E}{\partial \textrm{u}_s^{[k+1]}}$) that is already known starting from  the error of loss $E$ w.r.t. $\textrm{u}^{[K]}_s$ :
\begin{equation} \label{eq:gradient1}
\frac{\partial E}{\partial \textrm{u}^{[K]}_s} = \frac{2}{I} ( \textrm{u}^{[K]}_s - \overline{\textrm{x}}_s ).
\end{equation}

The expression of the gradients are provided in Appendix A.  The backward procedure is detailed in Algorithm~\ref{algo:backwardk}.

\begin{algorithm}[h]
	\small
	\caption{Backward procedure of ``Full learning''} \label{algo:backwardk}
	\BlankLine
	\KwIn{Set $D^{[k]}$, $b^{[k]}$, $\eta^{[k]}$, $k=\{1,\ldots K\}$ according to \eqref{eq:newstructure} and set $\gamma>0$}
	\KwData{Set $\textrm{u}^{[1]}_s = A^*\mathrm{z}_s \;\mbox{for every} \;s=\{1, \ldots, I\}$}
	\For{$\ell = 1, \ldots, T$}{
		Select one training sample $\textrm{u}^{[1]}_s$, calculate the gradient of loss w.r.t. $\textrm{u}^{[K]}_s = f_{\Theta_{[\ell]}}(\textrm{u}^{[1]}_s)$ according to Eq.~\eqref{eq:gradient1};\\
		\For{$k =K-1, \ldots, 1$}{
			Backpropagate the errors to $\textrm{c}^{[k]}_s $ and $\textrm{u}^{[k]}_s$ in the $[k]$ layer according to Eq.~\eqref{eq:derivac} and~\eqref{eq:derivauk}; \\
		}
		\For{$k =K, \ldots, 1$}{	
			Calculate the gradients of the loss w.r.t. $D^{[k]}$ and $b^{[k]}$ according to Eq.~\eqref{eq:gradientDk} and~\eqref{eq:gradientBk};\\
			Calculate the gradients of the loss w.r.t. $\tau^{[k]}$, $\sigma^{[k]}$ and $L^{[k]}$ according to Eq.~\eqref{eq:taugrad}, \eqref{eq:sigmagra} and~\eqref{eq:Lgrad}.
		}	
		Update $\tau_{[\ell+1]}^{[k]}$, $\sigma_{[\ell+1]}^{[k]}$ and $L_{[\ell+1]}^{[k]}$ for layers $k\in\{1,\ldots K\}$ by Eq.~\eqref{eq:update}.
	}
\end{algorithm} 

\subsection{Full versus Partial learning} \label{sec:partiallearning}

We propose two versions of our network: Full DeepPDNet and Partial DeepPDNet. Full DeepPDNet denotes the proposed supervised strategy described in Section~\ref{sec:DeepPDNet} relying on the learning of $\tau^{[k]}$, $L^{[k]}$, and $\sigma^{[k]}$. Partial DeepPDNet is reduced to the learning of $\tau^{[k]}$ and $L^{[k]}$ while $\sigma^{[k]}$ is set as:
\begin{equation}
\sigma^{[k]} = \frac{1/\tau^{[k]} - \Vert A \Vert^2/2}{\Vert L^{[k]} \Vert^2}.
\end{equation} 
This choice is guided by the technical assumptions insuring the convergence of the sequence $(\textrm{x}^{[k]})_{k\in \mathbb{N}}$, generated by standard Condat-V\~u algorithm  (cf. Algorithm~\ref{algo:primaldual}), to $\widehat{\mathrm x}_\lambda$ defined in \eqref{eq:basic_prob}. 
In the experimental section, we investigate the performance for both learning strategies (cf. Section~\ref{sec:experiment}).

\section{Global vs local structured $L$}
\label{s:gls}
The penalization term $g(L \textrm{x})$ in Eq.~\eqref{eq:basic_prob} can be considered as prior inforcing some smoothness on the solution. In many references of the recent inverse problem literature, $g$ denotes a $\ell_1$-norm or a $\ell_{1,2}$-norm in order to obtain sparse features. In the scenario of image restoration, $L$ usually models the discrete horizontal and vertical difference or other linear operator allowing to capture discontinuities. From other published research, it is well known that the structure of $L$ has a crucial impact on the performance of image restoration. 

\begin{figure}[t]
	\centering
	\includegraphics[width=0.3\linewidth]{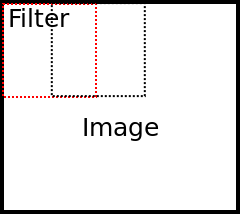}
	\caption{Sliding window modelling the location of the non-zero coefficients for each row of $L^{[k]}$ in the \textit{local features} setting.} \label{fig:localinitL}
\end{figure}

In the proposed DeepPDNet, each layer $k \in \{1,\ldots, K\}$ involves a learned linear transform $L^{[k]} \in \mathbb{R}^{P \times N}$,  where each row corresponds to a learned pattern of the image. 
In this work, we study three architectures for $L^{[k]}$: (i)~ $L^{[k]}$ is a dense matrix, (ii)~$L^{[k]}$ is a block-sparse matrix, (iii)~$L^{[k]}$ is built as a combination of dense and block-sparse matrices, respectively describing the global, the local, and the mixed global-local patterns. More specifically,
\begin{itemize}
	\item Global patterns: $L^{[k]}$ is built without any prior knowledge. It is a dense matrix  from the image space $\mathbb{R}^{N}$ to a feature space $\mathbb{R}^{P}$. Each row attempts to learn one type of global structure that occur in the image. In practice, $L^{[k]}$ is initialized by random values following a normal distribution.
	\item Local patterns: we construct a matrix $L^{[k]}$ with local sparse structures, which is inspired by the local patch dictionary~\cite{Boulangerinverse2018}. 
	For each row of $L^{[k]}$, $Q \times Q$ non-zero coefficients are locating in the region illustrated in Fig.~\ref{fig:localinitL}. The location of the window, and thus of the non-zero coefficients, is slided through the image.  In our experiments, the values of the non-zero coefficients are randomly initialized according to a normal distribution. In the learning procedure, the non-zeros elements in $L^{[k]}$ are updated, the other ones remain zero.
	\item Mixed global-local patterns: $L^{[k]}$ is built as a combination of dense and block-sparse matrices.
\end{itemize}

Global and local $L^{[k]}$ are respectively related to the fully connected layer and the convolution layer, but they are not the same.  Indeed, if we consider a network with $K$ layers formally defined as in \eqref{equa:reform},  the fully connected layer or convolutional layer are implemented in $(D^{[k]})_{1\leq k \leq K}$ as being either a dense matrix or a block-sparse matrix, while in our network, $D^{[k]}$ has a special structure coming from Condat-V\~u proximal iterations, whose expression is given in Section~\ref{sec:DeepPDNet} as:
$$
{\small{ D^{[k]} = \begin{pmatrix}
		\mathrm{Id} - \tau^{[k]} A^*A & - \tau^{[k]} (L^{[k]})^*\\
		\sigma^{[k]} L^{[k]}(\mathrm{Id} - 2\tau^{[k]} A^*A) &\mathrm{Id} - 2\tau^{[k]} \sigma^{[k]} L^{[k]} (L^{[k]})^* \\
		\end{pmatrix}.}}
$$
In this work, we consider similarly either dense matrix or block-sparse matrix but at the level of the analysis operator $L^{[k]}$.

Among the three configurations listed above (global, local-sparse, and mixed global-local), the combination of mixed global-local is the closest from recent CNNs, where fully connected layers and convolutional layers are combined in a cascade way. The numerical impact of these different choices  for $L^{[k]}$ will be discussed in the next section.

\section{Experiments} \label{sec:experiment}

\subsection{Network}

A deep primal dual network with $K$ layers has been built according to Section~\ref{sec:dpdn} with $g = \Vert \cdot \Vert_1$ and the weights in each layer are initialized by Eq.~\eqref{eq:newstructure} such that $\tau^{[k]}\equiv 1$, $L^{[k]} \in \mathbb{R}^{P\times N}$ (where $N$ is the raw image dimensionality and $P$ denotes the embedded feature number) is randomly initialized by values following a normal distribution with a standard deviation set to $10^{-2}$, and $\sigma^{[k]}=\frac{(1/\tau^{[k]} - \Vert A \Vert^2/2)}{\Vert L^{[k]}\Vert^2}$.

\subsection{Dataset}
\noindent We consider a training set $\mathcal{S} = \{ (\overline{\textrm{x}}_s, \textrm{z}_s) | s=1, \ldots, I\}$ and an evaluation set $\mathcal{T}=\{ (\overline{\textrm{x}}_s, \textrm{z}_s) | s=1, \ldots, J\}$ respectively containing $I$ images  and $J$ images, where $\overline{\textrm{x}}_s$ is the original image and $\textrm{z}_s$ is its degraded counterpart obtained by the degradation model~\eqref{eq:degrad_model}. 

Two different inverse problems are considered: image restoration and single image super-resolution. In image restoration experiments, $A$ models a uniform blur and the noise is a white Gaussian noise with variance $\alpha^2$. The size of the blur and the variance are specified for each experiment. For single image super-resolution, $A$ denotes a decimation operator and no noise is considered.

We evaluate the performance of the proposed network in the context of image restoration on MNIST dataset~\cite{LeCun98} and BSD68 dataset~\cite{Roth2009Fields} and in the context of super-resolution on BSD100 dataset~\cite{TimofteSmetaccv2014} and SET14~\cite{ReydeEladProtter2012} datasets used in~\cite{KimLeeLee_vdsr_cvpr16}.

\subsection{Performance evaluation}
\noindent The restoration performance are evaluated in terms of PSNR (i.e.~Peak Signal-to-Noise Ratio) and SSIM (i.e.~Structural SIMilarity), where higher values stand for better performance. 

The convergence of the primal-dual scheme on which relies the deep learning procedure is insured when condition~\eqref{eq:condpd} is satisfied. To estimate the distance to this constraint we compute: 
\begin{align} \label{eq:constraint}
d_C(\tau^{[k]},\sigma^{[k]}, L^{[k]})& = \max\Big(0,  \frac{\Vert A \Vert^2}{2} - \frac{1}{\tau^{[k]}} + \sigma^{[k]}  \Vert L^{[k]}\Vert^2 \Big)^2 
\end{align}

The simulations have been performed by a workstation with 4 cores and each is 3.20GHz (Intel Xeon(R) W-2104 CPU) and 64G memory. The code is implemented in MATLAB. 

\subsection{Performance on MNIST dataset} 
\label{ssec:mnist}
\noindent The MNIST dataset is a widely used handwritten benchmark for classification, containing 60,000 training images and 10,000 test images and each has a dimensionality of 28$\times$28. Instead of a classification task, in this work, it is used for image restoration. In the training procedure, the training set is further split into two subsets: 50,000 for training and the rest for validation. A hold-out validation scheme is applied to estimate the network architecture.

\begin{figure*}[t]
	\centering
	\includegraphics[width=0.8\linewidth]{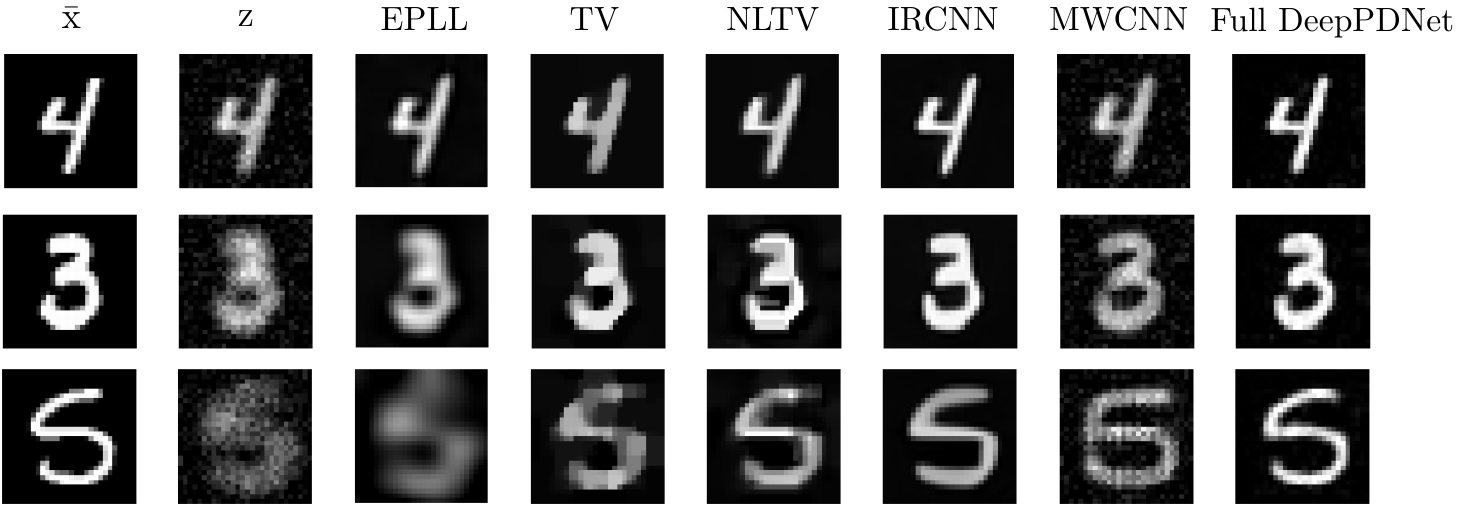}
	\caption{Visual comparisons on MNIST dataset for different methods. The first row corresponds to the MNIST data with a uniform $3\times3$ blur and a Gaussian noise with $\alpha=20$, the second row is with a uniform $5\times5$ blur and a Gaussian noise with $\alpha=20$, the third row is with a uniform $7\times7$ blur and a Gaussian noise with $\alpha=20$. For each instance, the images from the first to the seventh column respectively correspond to the original one $\bar{\textrm{x}}$, the degraded one $\textrm{z}$, the restored ones by EPLL, TV, NLTV, IRCNN and the proposed full DeepPDNet ($K=6$).} \label{fig:mnist-deblur-compar} 
\end{figure*}

The instances of clean images and their respective degraded ones are shown in Fig.~\ref{fig:mnist-deblur-compar}. In the following experiments, we adopt a mini-batch stochastic gradient descent algorithm to update the parameters with a batch size of 200 for the network learning. 
The maximum iteration is set to  $3\times10^4$.

\begin{figure*}[t]
	\centering
	\includegraphics[width=0.23\linewidth]{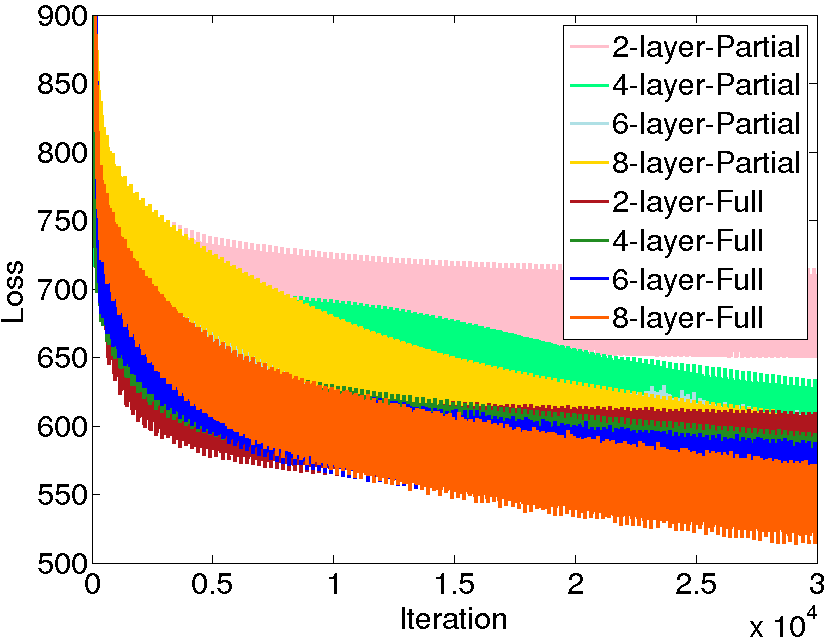}
	\includegraphics[width=0.23\linewidth]{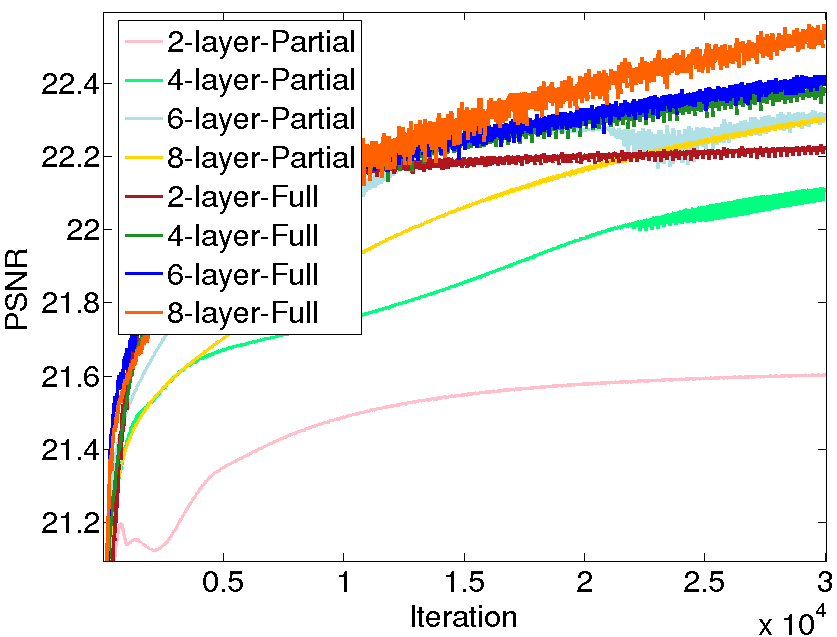} 
	\includegraphics[width=0.23\linewidth]{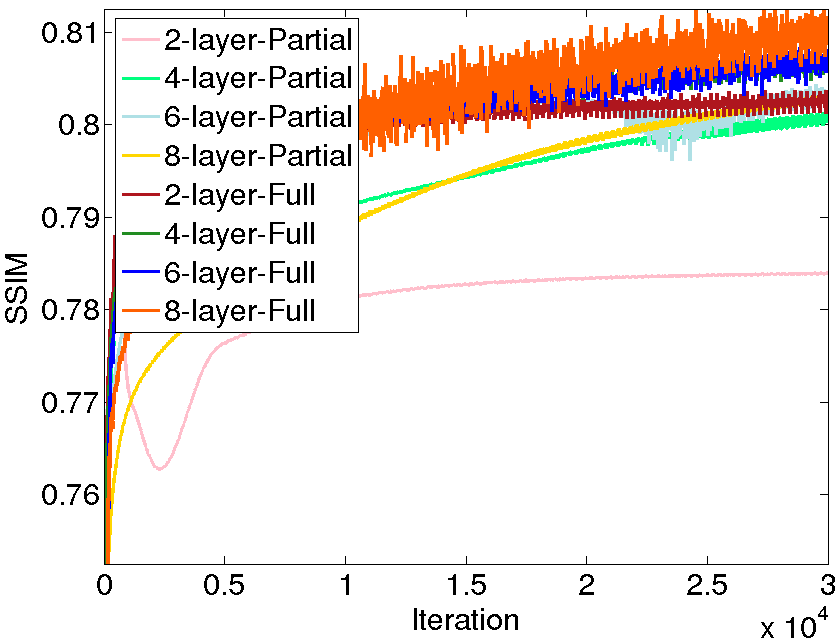}
	\includegraphics[width=0.23\linewidth]{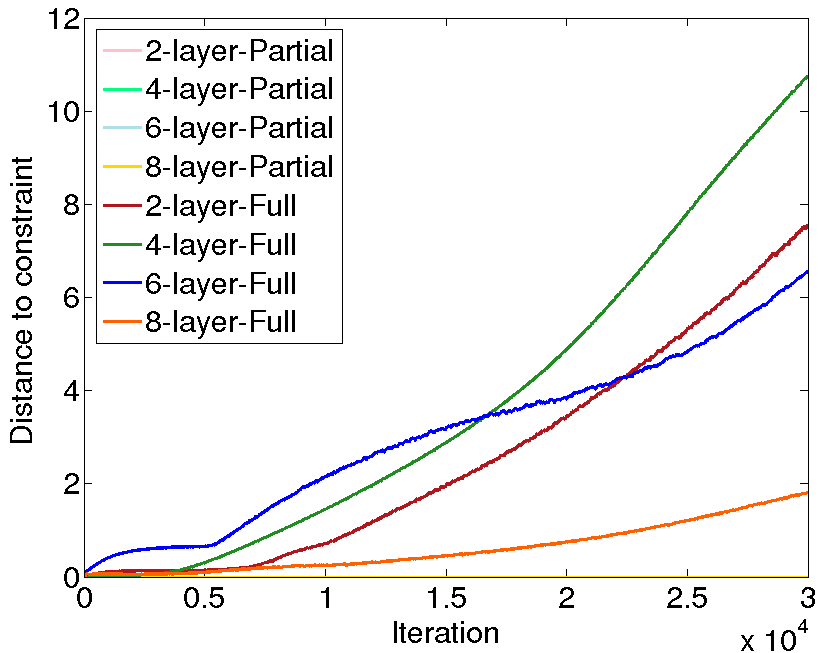}
	\caption{From left to right: Training loss, performance in terms of PSNR and SSIM on the validation set of MNIST dataset set, and $d_C$ defined in \eqref{eq:constraint} for full and partial learning with 2-layer, 4-layer, 6-layer and 8-layer networks and $P=100$. The results are obtained from data degraded with a uniform $3\times3$ blur and a Gaussian noise with $\alpha=20$. (Best visualisation in color)} \label{fig:mnistlayer}
\end{figure*}
\begin{figure*}[ht]
	\centering
	\includegraphics[width=0.23\linewidth]{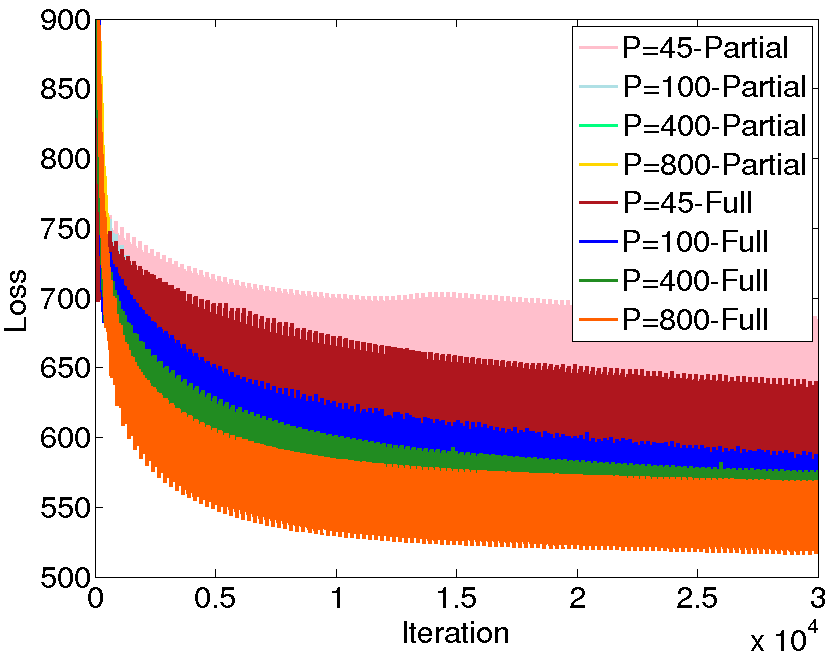} 
	\includegraphics[width=0.23\linewidth]{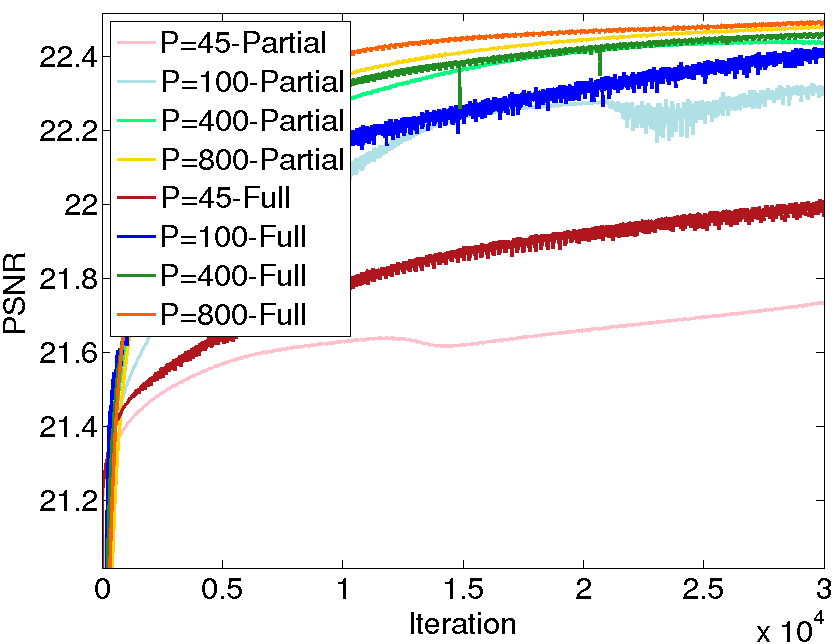} 
	\includegraphics[width=0.23\linewidth]{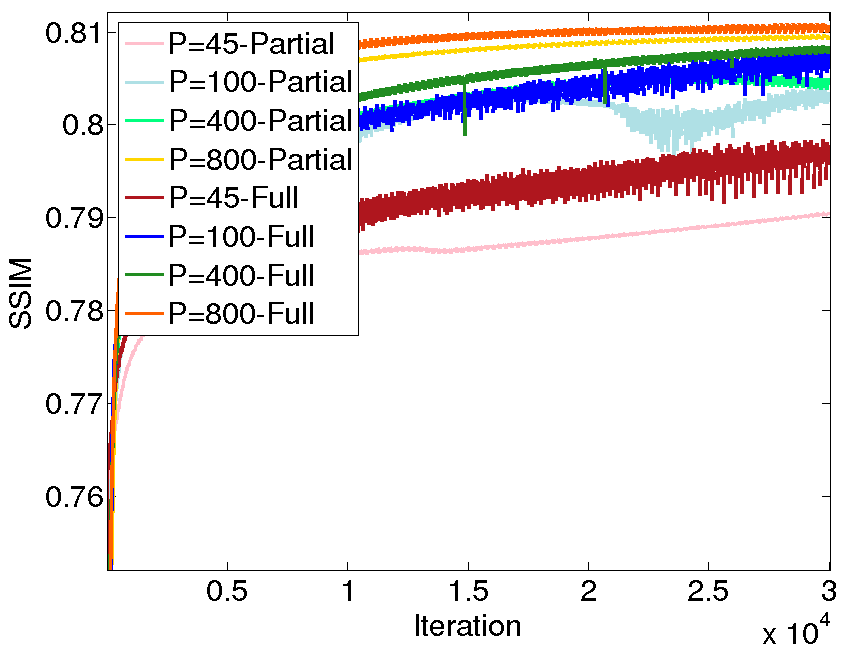}
	\includegraphics[width=0.22\linewidth]{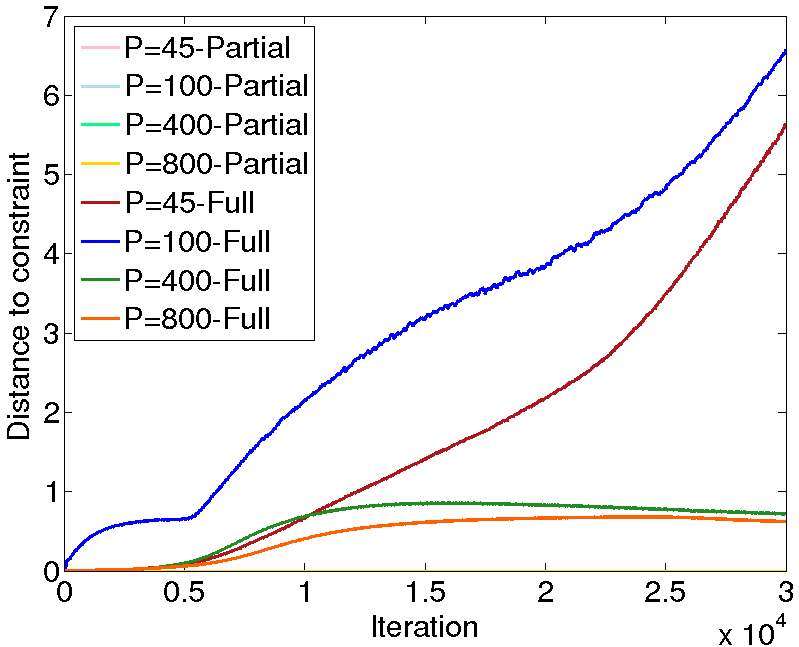}
	\caption{From left to right: Training loss, performance in terms of PSNR and SSIM on the validation set of MNIST dataset set, and $d_C$ defined in \eqref{eq:constraint} for full and partial learning with 6-layer and $P=25, 100, 400, 800$. The results are obtained from data degraded with a uniform $3\times3$ blur and a Gaussian noise with $\alpha=20$. (Best visualisation in color)} \label{fig:mnistP}
\end{figure*}

\noindent \textbf{Impact of network depth (i.e. $K$) -- }
To study the impact of the architecture depth of the network, we conduct the experiments for Partial DeepPDNet and Full DeepPDNet described in Section~\ref{sec:partiallearning} on 2-layer, 4-layer, 6-layer, and 8-layer networks with a fixed embedded feature number $P=100$ and  with global $L^{[k]}$. The simulations are performed with a uniform  $3\times3$ blur  and an additive Gaussian noise with a standard deviation $\alpha = 20$. The training loss, PSNR, SSIM, and the distance to convex constraint in the last layer w.r.t.~the iterations on the validation set are displayed in Fig.~\ref{fig:mnistlayer}. It can be seen that: i) Full DeepPDNet gives better results than Partial DeepPDNet; ii) Deeper architecture generally produces better results due to more meaningful feature learning; iii) the 8-layer network obtains a marginal gain compared to its 6-layer counterpart. The average cost time for one iteration of one mini-batch learning (including forward and backward process) is presented in Tab.~\ref{tab:timelayer} for different configurations of Full DeepPDNet. In our simulations, we perform training considering 30,000 iterations, which leads to a learning time going from 2 hours to 13 hours depending on the number of layers. Combining the computation cost and that marginal benefit with the 8-layer network, we adopt the architecture of a 6-layer network in the following experiments.

\noindent \textbf{Impact of the size of $L^{[k]}$ (i.e. $P$) --} For the 6-layer network, we study the impact of the size of $P$ (i.e.~$P=25, 100, 400, 800$) in the scenario of Partial DeepPDNet and Full DeepPDNet described in Section~\ref{sec:partiallearning}  with global $L^{[k]}$. The simulations are performed with a uniform  $3\times3$ blur  and an additive Gaussian noise with a standard deviation $\alpha = 20$. The results are displayed in Fig.~\ref{fig:mnistP}. It can be seen that: i) Full DeepPDNet leads to better performance than Partial DeepPDNet for similar $P$; ii) Larger $P$ leads to better results due to more feature embedding. However, when $P=800$, the performance is close to the ones with $P=400$, which demonstrates that when $P$ becomes large, the capacity of improvement becomes limited.

\noindent \textbf{Full versus Partial learning --} From the above results, we can observe that the learning capacity is boosted when the feasible parameter space is enlarged (cf. with Full DeepPDNet),  leading to a better solution than with Partial DeepPDNet. 

The difference between Partial DeepPDNet and Full DeepPDNet only relies on the estimation of $K$ additional parameters $(\sigma^{[k]})_{1\leq k \leq K}$, which is small compared to the $K\times (PN+1)$ parameters due to the learning of $(L^{[k]}, \tau^{[k]})_{1\leq k \leq K}$ in the context of dense matrices or even compared to the $K\times (PQ^2+1)$ when block-sparse matrices are involved. This is certainly the reason why we do not observe overfitting. Indeed,  from the results displayed in Fig.~\ref{fig:mnistlayer} and Fig.~\ref{fig:mnistP}, we can observe that, for each configuration, the training loss decreases as a function of the iterations while the PSNR and SSIM on the validation set increases w.r.t. the iterations and does not imply performance drop. This quantitative results illustrate that no overfitting appears either with or without additional learned parameter $\sigma^{[k]}$.

\begin{table*}[t]
	\resizebox{\textwidth}{!}{
		\begin{tabular}{|c|c|c|c|c|c|c|c|c|c|c|}
			\hline
			Setting & ``f28s28n10" & ``f14s14n10" & ``f14s7n10" & ``f9s9n10" & ``f9s4n10" & ``f7s7n10" & ``f7s3n10" & ``f5s5n10" & ``f5s2n10" & ``f3s3n10" \\
			\hline
			P & 10 & 40 & 90 & 90 & 160 & 160 & 640 & 250 & 1210 & 810 \\
			Sparsity rate & 0\% & 75\% & 75\% & 89.67\% & 89.67\% & 93.75\% & 93.75\% & 96.81\% & 96.81\% & 98.85\%\\
			\hline
	\end{tabular}}
	\caption{Value of $P$ and sparsity rate for different choices of local sparse $L^{[k]}$. The smaller is the sparsity rate, the denser is $L^{[k]}$. } \label{tab:embeddingPnumber}
\end{table*}

\begin{table}[h]
	\centering
	\begin{tabular}{|c|c|}
		\hline
		Architecture & Time (\textrm{s}) \\
		\hline
		2-layer & 0.3460  \\
		4-layer & 0.8368 \\
		6-layer & 1.2767 \\
		8-layer & 1.6084 \\
		\hline
	\end{tabular}
	\caption{Average cost time (in \textrm{s}) for one learning iteration (including forward and backward) of one mini-batch of MNIST dataset for different network configurations.\label{tab:timelayer}}	
	\vspace{-0.3cm}
\end{table}

\noindent \textbf{Distance to the constraint -- }
Here, we investigate the distance to the convex set for the full learning in two different viewpoints: the distance for different depths and also different $P$ in the last layer of the networks. The distances to the constraint~\eqref{eq:constraint} w.r.t. the iterations for different depth and different $P$ are shown in the last column of Fig.~\ref{fig:mnistlayer} and Fig.~\ref{fig:mnistP} respectively. It can be seen that the distance in the last layer decreases as the network becomes deep and also as $P$ value becomes large, which reflects the learning ability of a large network. From these results, although full learning relaxes the constraint between the parameters, it has the ability to make the violation distance smaller when making the network deeper or larger. Obviously, the constraint is always satisfied with Partial learning.  

\begin{table}[tbp]
	\centering
	\begin{tabular}{|c|cc|cc|}
		\hline
		& \multicolumn{2}{c|}{PSNR} & \multicolumn{2}{c|}{SSIM} \\
		\hline
		$P$ & Global & Local sparse & Global & Local sparse \\
		\hline
		10 & 21.64 & 21.61 & 0.7846 & 0.7831 \\
		40 & 22.23 & 22.22 & 0.8033 & 0.8041 \\
		90 & 22.35 & 23.06 & 0.8052 & 0.8287 \\
		160 & 22.35 & 23.06 & 0.8052 & 0.8370 \\
		250 & 22.40 & 22.72 & 0.8059 & 0.8466 \\
		640 & 22.49 & 24.48 & 0.8076 & 0.9122 \\
		810 & 22.50 & 24.37 & 0.8113 & 0.9202 \\
		1210 & 22.49 & 24.80 & 0.8112 & 0.9278 \\
		\hline
	\end{tabular}
	\caption{Comparison results of global and local $L^{[k]}$ according to Tab.~\ref{tab:embeddingPnumber} on the validation set of MNIST dataset from data degraded by a uniform $3\times3$ blur and a Gaussian noise with $\alpha=20$. $P=90$ corresponds to the setting of ``f14s7n10'' showing better performance than ``f9s9n10'' and $p=160$ corresponds to the setting of ``f7s7n10''. } \label{tab:mnistglobalvslocalcomp}
\end{table}

\begin{table}[tbp]
	\centering
	\begin{tabular}{|c|c|c|}
		\hline
		Fusion & P & PSNR/SSIM \\
		\hline
		``f5s2n10'' & 1210 & 24.80/0.9278   \\
		``f5s2n10''+``f7s3n10'' & 1850 & 25.04/0.9317 \\
		``f5s2n10''+``f7s3n10''+``f14s7n10'' & 1940  &  25.06/0.9301 \\
		``f5s2n10''+``f7s3n10''+``f14s7n10''+``f28s28n10''& 1950 & 25.33/0.9335 \\
		\hline
	\end{tabular}
	\caption{Performance of combination of multiple local sparse filters on the validation MNIST dataset with uniform blur filter $3\times3$ and Gaussian noise $\alpha=20$.\label{tab:mnistfusionperf}}	
\end{table}

\begin{table*}[tbp]
	\centering
	\resizebox{\textwidth}{!}{
	\begin{tabular}{|c|c|c|c|c|c|c|}
		\hline
		\multirow{3}{*}{Data} & \multirow{3}{*}{Method} & \multicolumn{3}{c|}{$3\times3$ Blur} & $5\times5$ Blur & $7\times7$ Blur  \\
		\cline{3-7}
		& & $\alpha=10$ & $\alpha=20$ & $\alpha=30$ & $\alpha=20$ & $\alpha=20$ \\
		\cline{3-7}
		& & PSNR/SSIM & PSNR/SSIM & PSNR/SSIM & PSNR/SSIM & PSNR/SSIM\\     
		\hline
		\multirow{7}{*}{MNIST} & EPLL~\cite{Zoran_D_2017_p-iccv_fro_lmn} & 24.02/0.8564 & 20.99/0.7628 & 19.05/0.6871 & 16.42/0.5629 & 13.97/0.3265 \\
		& TV~\cite{RudinOsher92PhysD} & 25.07/0.8583 & 19.58/0.7004 & 18.86/0.6681 & 18.86/0.6681 & 16.31/0.5665 \\
		& NLTV~\cite{ChierchiaNelly2014tip} & 25.49/0.8697 & 21.98/0.7738 & 20.73/0.7353 & 20.73/0.7353 & 16.79/0.6228 \\
		& MWCNN~\cite{LieZhang-wmcnn_cvpr18}  & 19.16/0.7219 & 18.53/0.6782 & 17.78/0.6499 & 15.83/0.5343 & 13.04/0.3175 \\	
		& IRCNN~\cite{ZhangZuo2017ircnn} & \textbf{28.52}/0.8904 & 25.00/0.8193 & 22.63/0.7723 & 21.46/0.7698 & 18.29/0.6546 \\			
		& Partial DeepPDNet & 23.67/0.8366 & 22.03/0.7983 & 20.93/0.7750 & 17.96/0.6534 & 16.21/0.5505\\    
		& Full DeepPDNet & {27.40}/\textbf{0.9410} & \textbf{25.09}/\textbf{0.9254} & \textbf{23.61}/\textbf{0.9097} & \textbf{22.43}/\textbf{0.8738} & \textbf{20.43}/\textbf{0.8157} \\
		\hline
	\end{tabular}
	}
	\caption{Comparison results of different methods on the MNIST dataset from different degradation configurations.} \label{tab:mnistfinalcomp}
\end{table*}

\begin{figure*}[tbp]
	\centering
	\includegraphics[width=0.25\linewidth]{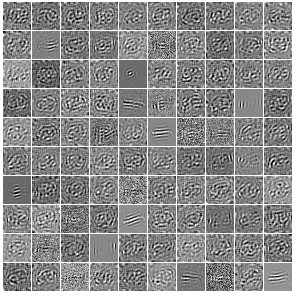}
	\includegraphics[width=0.28\linewidth]{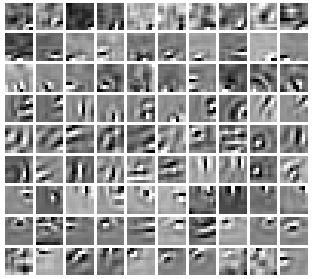}
	\caption{Visualization of the rows of $L^{[k]}$ for the layer $k=6$ when considering  Full DeepPDNet in the context of data degraded by a uniform $3\times3$ blur and an additive Gaussian noise with $\alpha=20$ on MNIST dataset. (a)~Global patterns when $P=100$: each image is associated with a reshaped row of $L^{[6]}$  leading to 100 small images of size $\sqrt{N}\times \sqrt{N}$. (b)~Reshaped rows of the block-sparse $L^{[6]}$ for the setting ``f9s9n10''  when $Q=9$, $P=90$ leading to 90 small filters of size $9\times 9$.}\label{fig:mnistlearnedpattern}
\end{figure*}

\begin{table*}[tbp]
	\centering
	\resizebox{\textwidth}{!}{
	\begin{tabular}{|c|c|c|c|c|c|c|c|c|}
		\hline
		\multirow{3}{*}{Data} & \multirow{3}{*}{Noise} & \multirow{3}{*}{Method} & \multicolumn{2}{c|}{$3\times3$ Blur} & \multicolumn{2}{c|}{$5\times5$ Blur} & \multicolumn{2}{c|}{$7\times7$ Blur}  \\
		\cline{4-9}
		& & & \multicolumn{2}{c|}{$\alpha=20$} & \multicolumn{2}{c|}{$\alpha=20$} & \multicolumn{2}{c|}{$\alpha=20$}\\
		\cline{4-9}
		& & & PSNR/SSIM & Relative Drop & PSNR/SSIM & Relative Drop & PSNR/SSIM & Relative Drop \\     
		\hline
		\multirow{12}{*}{MNIST} & \multirow{3}{*}{$\beta=2$} & MWCNN & 18.52/0.6763 & 0.05/0.28 & 15.83/0.5340 & 0/0.06 & 13.04/0.3175 & 0/0 \\
		& & IRCNN & 24.84/0.8149 & 0.64/0.54 & 21.39/0.7654 & 0.33/0.57 & 18.24/0.6501 & 0.27/0.69 \\
		& & Full DeepPDNet & 25.05/0.9226 & 0.16/0.30 & 22.40/0.8709 & 0.13/0.33 & 20.40/0.8127 & 0.15/0.37 \\
		\cline{2-9}
		& \multirow{3}{*}{$\beta=5$} & MWCNN & 18.46/0.6722 & 0.38/0.88 & 15.80/0.5320 & 0.19/0.43 & 13.01/0.3158 & 0.23/0.54 \\
		& & IRCNN & 24.61/0.8079 & 1.56/1.39 & 21.24/0.7584 & 1.03/1.48 & 18.14/0.6428 & 0.82/1.80 \\
		& & Full DeepPDNet & 24.90/0.9159 & 0.76/1.03 & 22.29/0.8640 & 0.62/1.12 & 20.29/0.8051 & 0.69/1.30 \\
		\cline{2-9}		
		& \multirow{3}{*}{$\beta=10$} & MWCNN & 18.26/0.6639 & 1.46/2.11 & 15.67/0.5257 & 1.01/1.61 & 12.85/0.3068 & 1.46/3.37 \\
		& & IRCNN & 24.04/0.7958 & 3.84/2.87 & 20.89/0.7459 & 2.66/3.10 & 17.90/0.6294 & 2.13/3.85  \\ 
		& & Full DeepPDNet & 24.45/0.8994 & 2.55/2.81 & 21.95/0.8465 & 2.14/3.12 & 19.93/0.7852 & 2.45/3.74 \\	
		\cline{2-9}		
		& \multirow{3}{*}{$\beta=20$} & MWCNN & 17.56/0.6433 & 5.23/5.15 & 15.07/0.4968 & 4.80/7.02 & 12.22/0.2688 & 6.29/15.34 \\
		& & IRCNN & 22.51/0.7693 & 9.96/6.10 & 19.84/0.7157 & 7.55/7.03 & 17.17/0.5966 & 6.12/8.86 \\
		& & Full DeepPDNet & 22.99/0.8547 & 8.37/7.67 & 20.77/0.7954 & 7.40/8.97 & 18.70/0.7224 & 8.47/11.44 \\		
		\hline
	\end{tabular}
	}
	\caption{Comparisons of PSNR, SSIM and relative drops (in $\%$) on the test MNIST dataset degraded with noise to evaluate the robustness when considering MWCNN, IRCNN and the proposed ``Full DeepPDNet''. } \label{tab:mnistrobustperfcomp}
\end{table*}

\begin{figure*}[t]
	\centering
	\includegraphics[width=0.9\linewidth]{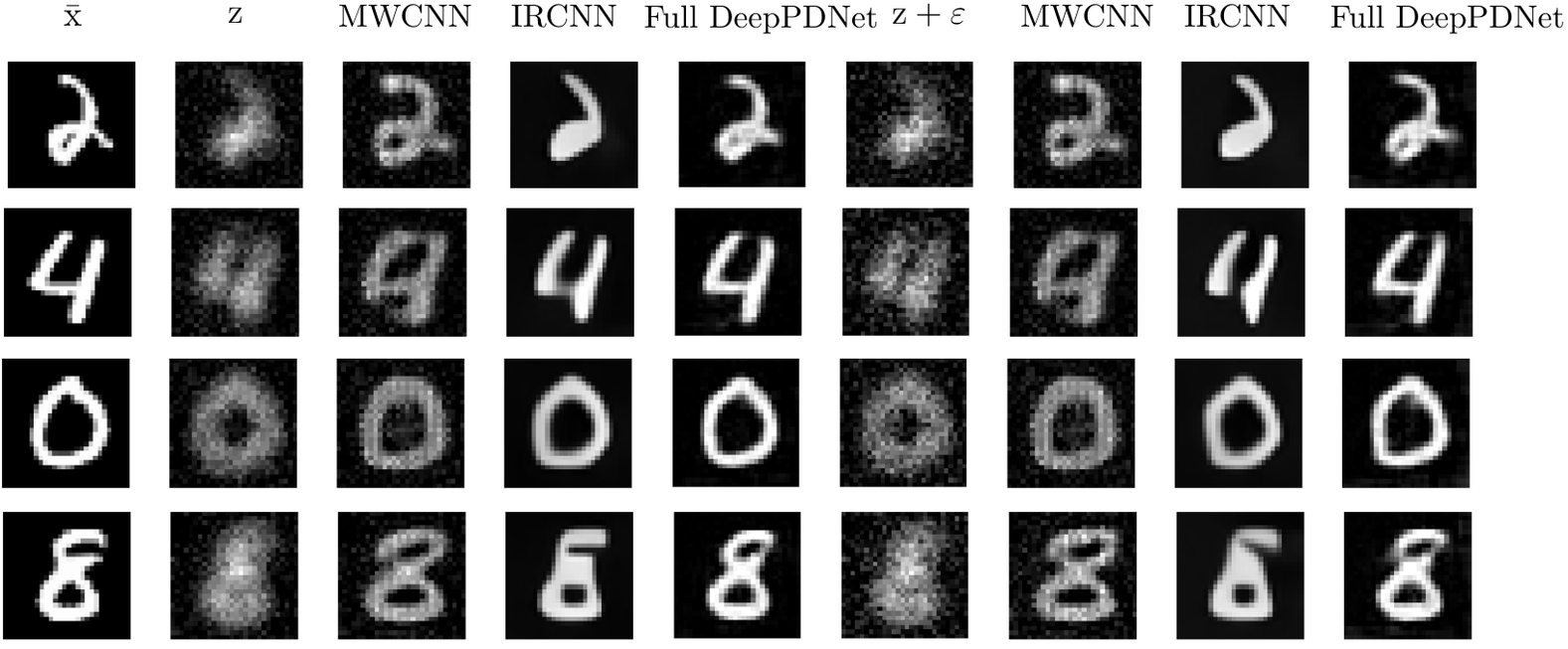}
	\caption{Visual comparisons of robustness analysis on the MNIST dataset with a uniform $7\times7$ blur and a Gaussian noise with $\alpha=20$. From the first row to fourth row, the addition noises are $\beta=2$, $\beta=5$, $\beta=10$ and $\beta=20$. For each row, the images are the clean image, the blur image, the result by MWCNN, the result by IRCNN, the result by the learned full DeepPDNet ($K=6$), the blur image with additional noise and its result by MWCNN, IRCNN, and the learned full DeepPDNet ($K=6$). } \label{fig:mnist-robust-analysis} 
\end{figure*}

\noindent \textbf{Local vs global matrix -- } 
Next, we investigate the performance of dense matrix (global) and block-sparse matrix (local) described in Section~\ref{s:gls}. We consider ten types of local sparse projection in $L$ for the 6-layer network: ``f28s28n10", ``f14s14n10", ``f14s7n10", ``f9s9n10", ``f9s4n10", ``f7s7n10", ``f7s3n10", ``f5s5n10", ``f5s2n10" and ``f3s3n10", which is named as the format of ``fQsNnS" such that ``Q" stands for the size of local square filter, ``N" means the strip length between two neighboring filters and ``S" means the number of filters at the same location. The corresponding $P$ number for each type is respectively shown in Tab.~\ref{tab:embeddingPnumber}, at the same time, we also give the sparsity rate of $L$. It is noted that local filter ``f28s28n10'' with $P=10$ corresponds to a dense matrix. To provide fair comparisons of the performance between global projection and local sparse projection, for each type of local sparse projection, we create a global projection with the same value for $P$. Their performance on the validation set is shown in Tab.~\ref{tab:mnistglobalvslocalcomp}. From the results, it is seen that: i) when $P$ becomes larger,  the performance of global projection slightly increase and remain stable when $P=810$; ii) The performance of local sparse projection obtain significant improvement compared to global projection when the same  $P$ is adopted, especially for large $P$; iii) The performance of global matrix converges earlier than local sparse projection as $P$ increases.

\noindent \textbf{Fusion of multiple block-sparse matrices:} Previously, we observe that block-sparse (i.e. local) $L^{[k]}$ has better performance than dense (i.e. global) ones. Since the block-sparse configuration is able to learn meaningful local patterns (as shown in Fig.~\ref{fig:mnistlearnedpattern}), then we propose to investigate the performance of the fusion of multiple of  block-sparse matrices.  The performance of a combination of  block-sparse matrices on the validation set is shown in Tab.~\ref{tab:mnistfusionperf}. We can conclude that the performance is further boosted when multiple block-sparse matrices are fused.

\noindent \textbf{Evaluation and comparisons -- } 
From the previous set of experiments, we choose to focus our test using Full DeepPDNet with  6 layers and fusion of multiple block-sparse matrices ``f5s2n10'', ``f7s3n10'' and ``f14s7n10'' as well as global projection ``f28s28n10''. We evaluate the performance on the test set for five simulations considering  different sizes of uniform blur (i.e.~$3\times3$, $5\times5$ and $7\times7$) with Gaussian noises of different standard deviation level (i.e~$\alpha =$ 10, 20, 30). 

Our method is compared to unsupervised strategy when $g(L\cdot)$ models either a TV penalization or an NLTV penalization~\cite{ChierchiaNelly2014tip}. The algorithmic procedure to estimate $\widehat{\textrm{x}}_\lambda$ relies on the iterations described in Algorithm~\ref{algo:primaldual} when the error between two successive updates are less than $10^{-5}$. We also compare to EPLL \cite{Zoran_D_2017_p-iccv_fro_lmn} which makes use of  the Gaussian mixture model to learn the prior for image deblurring. Comparisons to supervised CNN-based procedure: MWCNN~\cite{LieZhang-wmcnn_cvpr18} and IRCNN~\cite{ZhangZuo2017ircnn} are also provided~\footnote{We download the source codes from the authors' websites.}. MWCNN integrates  wavelet transform and inverse wavelet transform into U-Net architecture to contract the network and IRCNN employs a denoiser prior based on CNNs as a modular in the half quadratic splitting method for image restoration. The results on the test set are shown in Tab.~\ref{tab:mnistfinalcomp}. From the table, it is seen that the proposed DeepPDNet with Partial learning shows poor performance because of expressive ability limitation while full learning outperforms other methods significantly. We also give comparison results of average time cost per test image without using GPU for different methods in Tab.~\ref{tab:testtimecomp}, it is observed that the proposed DeepPDNet is faster than other methods, which is certainly due to the simplicity and the small number of layers in our framework.

\begin{table}[tbp]
	\centering
	\begin{tabular}{|c|c|}
		\hline
		Method & Time (\textrm{s}) \\
		\hline
		EPLL~\cite{Zoran_D_2017_p-iccv_fro_lmn} &  1.27  \\
		TV~\cite{RudinOsher92PhysD} &  0.07 \\
		NLTV~\cite{ChierchiaNelly2014tip} & 0.19 \\
		MWCNN~\cite{LieZhang-wmcnn_cvpr18} & 0.06 \\
		IRCNN~\cite{ZhangZuo2017ircnn} & 0.66 \\
		Full DeepPDNet & 0.01 \\
		\hline
	\end{tabular}
	\caption{Average cost time (in \textrm{s}) per test image for MNIST dataset for different methods.}\label{tab:testtimecomp}	
\end{table}

\begin{table*}[tbp]
	\centering
	\resizebox{\textwidth}{!}{
	\begin{tabular}{|c|c|c|c|c|c|c|c|}
		\hline
		\multirow{3}{*}{Data} & \multirow{3}{*}{Method} & \multicolumn{3}{c|}{Blur filter $3\times3$} & \multicolumn{3}{c|}{Blur filter $5\times5$}  \\
		\cline{3-8}
		& & $\alpha=15$ & $\alpha=25$ & $\alpha=50$ & $\alpha=15$ & $\alpha=25$ & $\alpha=50$ \\
		& & PSNR/SSIM & PSNR/SSIM & PSNR/SSIM & PSNR/SSIM & PSNR/SSIM & PSNR/SSIM \\     
		\hline
		\multirow{5}{*}{BSD68} & TV~\cite{RudinOsher92PhysD} & 25.52/0.6746 & 25.16/0.6634 & 23.27/0.5836 & 24.04/0.6141 & 23.83/0.6047 & 22.77/0.5622 \\
		& NLTV~\cite{ChierchiaNelly2014tip} & 25.86/0.6875 & 25.49/0.6780 & 23.52/0.5932 & 24.22/0.6238 & 24.02/0.6165 & 22.88/0.5711 \\
		& EPLL~\cite{Zoran_D_2017_p-iccv_fro_lmn} & 27.01/0.7450 & 25.60/0.6785 & 23.72/0.6137 & 25.32/0.6674 & 24.38/0.6198 & 22.99/0.5715 \\
		& MWCNN~\cite{LieZhang-wmcnn_cvpr18} & 26.56/0.7537 & 25.76/0.7136 & \textbf{23.88}/\textbf{0.6265} & 24.39/0.6533 & 24.03/0.6313 & 22.88/\textbf{0.5759} \\
		& IRCNN~\cite{ZhangZuo2017ircnn} & 26.78/\textbf{0.7840} & \textbf{26.13}/\textbf{0.7203} & 23.63/0.5981 & 24.66/\textbf{0.6947} & 24.64/\textbf{0.6555} & 22.96/0.5651 \\
		& Full DeepPDNet (Q=28, K=6) & \textcolor{black}{25.83/0.6628} & \textcolor{black}{24.63/0.6042} & \textcolor{black}{23.37/0.5789} & \textcolor{black}{24.44/0.6086} & \textcolor{black}{23.62/0.5612} & \textcolor{black}{22.27/0.5026} \\
		& Full DeepPDNet (Q=10, K=20) & \textbf{27.33}/0.7637 & 25.95/0.7055 & 23.69/0.6052 & \textbf{25.48}/0.6819 & \textbf{24.66}/0.6430 & \textbf{23.04}/0.5717 \\	   
		\hline
	\end{tabular}
	}
	\caption{Comparison results of different methods on the BSD68 dataset from different degradation configurations.} \label{tab:bsdfinalcomp}
\end{table*}

\begin{figure*}[tbp]
	\centering
	\begin{tabular}{cc}
		\includegraphics[width=0.35\linewidth]{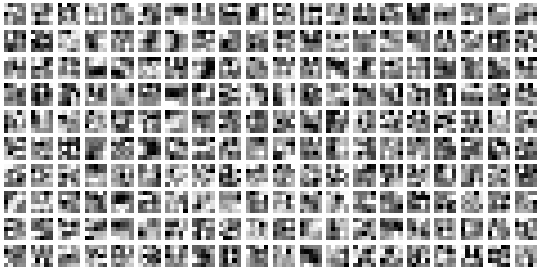}&
		\includegraphics[width=0.35\linewidth]{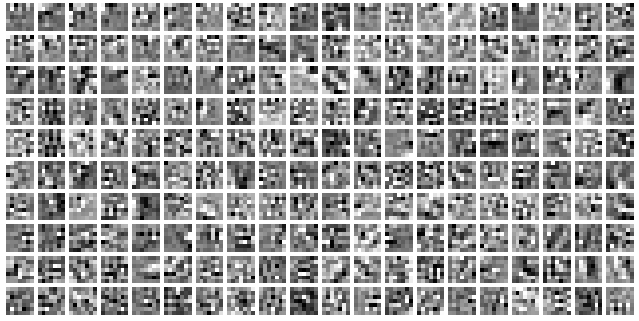}	\\
		(a) & (b)
	\end{tabular}
	\caption{Visualization of the non-zeros coefficients in $L^{[k]}$ for the layer $k=6$ when considering  Full DeepPDNet in the context of data degraded by a uniform $3\times3$ blur and an additive Gaussian noise with $\alpha=20$ on BSD68 dataset. (a)~Part of reshaped rows of the local sparse $L^{[6]}$ for the setting ``f5s2n10'' leading to $P=1210$. (b)~Part of reshaped rows of the local sparse $L^{[6]}$ for the setting ``f7s3n10'' leading to $P=490$. \label{fig:bsdlearnedpattern}}
\end{figure*}

Figure~\ref{fig:mnistlearnedpattern} and Figure~\ref{fig:bsdlearnedpattern} respectively display the learned $L^{[k]}$ for the layer $k=6$ on the MNIST and BSD400 dataset. In Figure~\ref{fig:mnistlearnedpattern}~(a), we display the results when $L^{[k]}$ denotes a dense matrix. The illustration is composed of 100 small images corresponding to the $P=100$ rows of $L^{[6]}$, each being reshaped as an image of size $\sqrt{N}\times \sqrt{N}$. Similarly, in  Figure~\ref{fig:mnistlearnedpattern}~(b), we display the learned rows of  $L^{[k]}$ when the local-sparse matrix is considered $Q=9$ and $P=90$, leading to 90 small images of size $Q\times Q$. Similar information is displayed in Figures~\ref{fig:bsdlearnedpattern}~(a) and (b) where both are associated with local-sparse  $L^{[k]}$ but considering either a construction with $Q=5$ (with the setting ``f5s2n10'')  or $Q=7$ (with the setting ``f7s3n10'').\\
\noindent \textbf{Robustness to noise}: We study the robustness of the proposed algorithm to additional noise. We consider different level of Gaussian noise of standard deviation $\beta$ to the test degraded images and forward them in the learned networks to obtain the restored ones. We evaluate the robustness for the learned models for different blur sizes: $3\times3$, $5\times5$, and $7\times7$ and Gaussian noise $\alpha=20$. In Tab.~\ref{tab:mnistrobustperfcomp}, we show the performance of the restored images and the relative performance decrease ratio compared to the results without additional noise in Tab.~\ref{tab:mnistfinalcomp}. The results of MWCNN and IRCNN are also given in Tab.~\ref{tab:mnistrobustperfcomp}. It can be observed that: i) when the additional noise has a small variance, the performance drop for the proposed method is lower than the one with IRCNN; ii) the relative performance drop for MWCNN is lower than the one with the proposed Full DeePDNet in most of scenarios, the possible reason is that wavelet transform is employed in the architecture to reduce the effect of additional noises; iii) when the additional noise is strong enough compared to the original noise, although serious performance drop occurs (for instance, $7\times7$ Blur when $\alpha=20$ and $\beta=20$) compared to MWCNN and IRCNN, the performance of our method are still better. To summarize,  it is observed that the relative performance drop ratio of the proposed method is sometimes higher than the drop with IRCNN and MWCNN, however, the performance are still better than with these two state-of-the-art CNN-based methods. Finally, we visualize some instances of the restored images by MWCNN, IRCNN and Full DeepPDNet without/with additional noises in Fig.~\ref{fig:mnist-robust-analysis}.

\begin{figure*}[tbp]
	\centering
	\begin{tabular}{llllllll}
		{\large \;\;\;\;\;\;\;$\bar{\textrm{x}}$} & {\large \;\;\;\;\;\;\;$\textrm{z}$} & {\small \;\;\;\;\;\;\;TV} & {\small \;\;\;\;\;\;NLTV} & {\small \;\;\;\;\;\;EPLL} & {\small \;\;\;\;MWCNN} & {\small \;\;\;\; IRCNN} & {\small Full DeepPDNet} \\
		\includegraphics[height=0.07\linewidth]{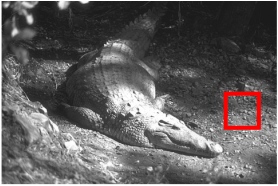} &
		\includegraphics[height=0.07\linewidth]{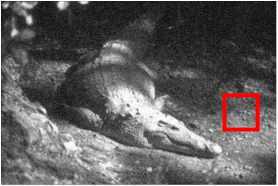} &
		\includegraphics[height=0.07\linewidth]{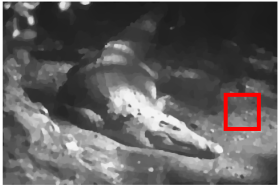} &
		\includegraphics[height=0.07\linewidth]{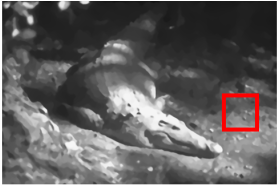} & 
		\includegraphics[height=0.07\linewidth]{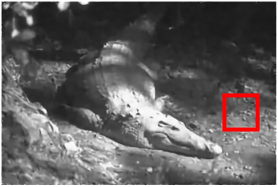} &
		\includegraphics[height=0.07\linewidth]{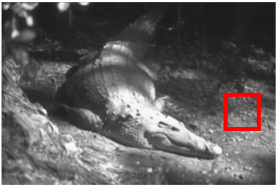} &
		\includegraphics[height=0.07\linewidth]{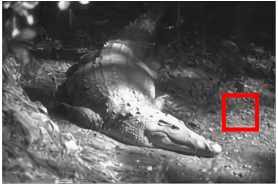} &
		\includegraphics[height=0.07\linewidth]{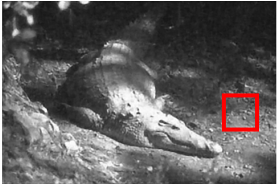} \\
		\includegraphics[height=0.05\linewidth]{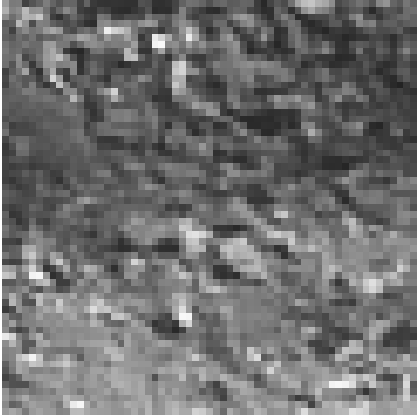} &
		\includegraphics[height=0.05\linewidth]{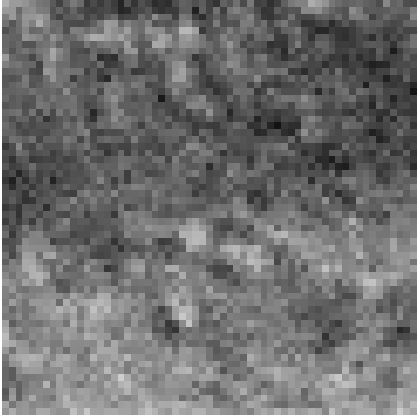} &
		\includegraphics[height=0.05\linewidth]{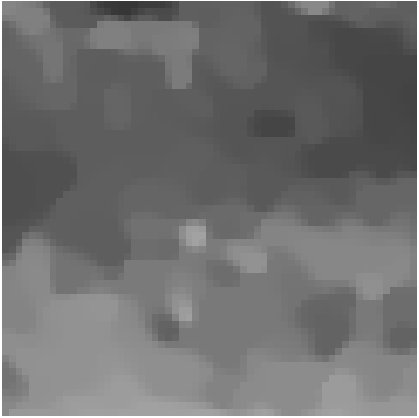} &
		\includegraphics[height=0.05\linewidth]{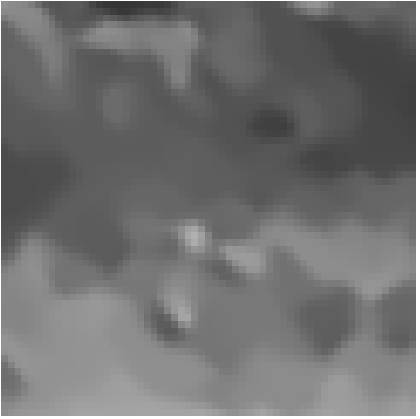} & 
		\includegraphics[height=0.05\linewidth]{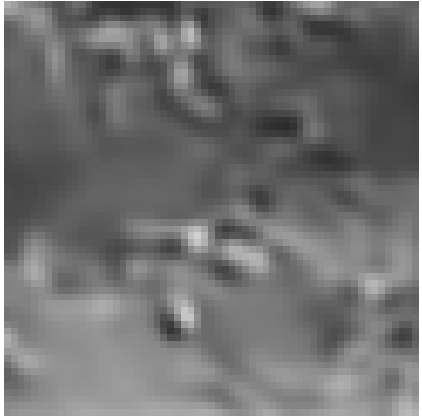} &
		\includegraphics[height=0.05\linewidth]{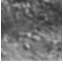} &	
		\includegraphics[height=0.05\linewidth]{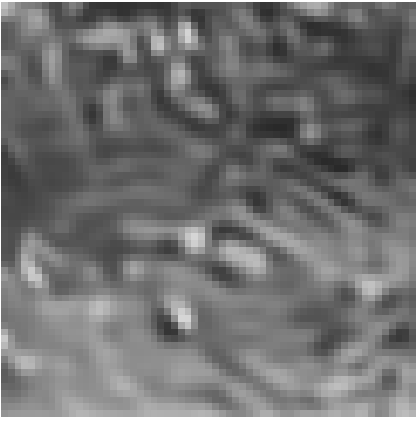} &
		\includegraphics[height=0.05\linewidth]{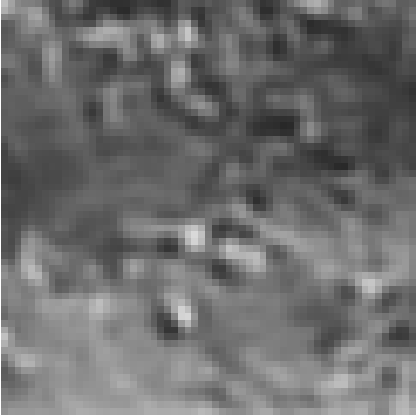} \\
		\includegraphics[height=0.07\linewidth]{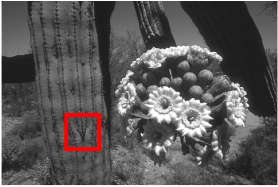} &
		\includegraphics[height=0.07\linewidth]{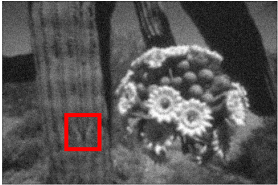} &
		\includegraphics[height=0.07\linewidth]{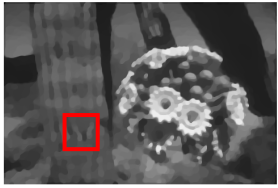} &
		\includegraphics[height=0.07\linewidth]{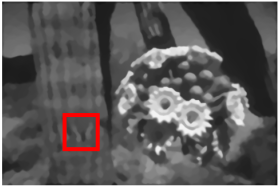} & 
		\includegraphics[height=0.07\linewidth]{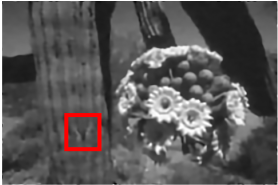} &
		\includegraphics[height=0.07\linewidth]{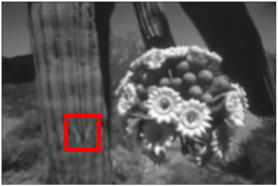} &	
		\includegraphics[height=0.07\linewidth]{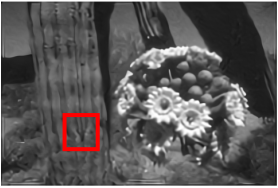} &
		\includegraphics[height=0.07\linewidth]{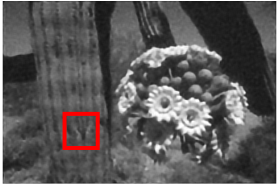}	\\
		\includegraphics[height=0.05\linewidth]{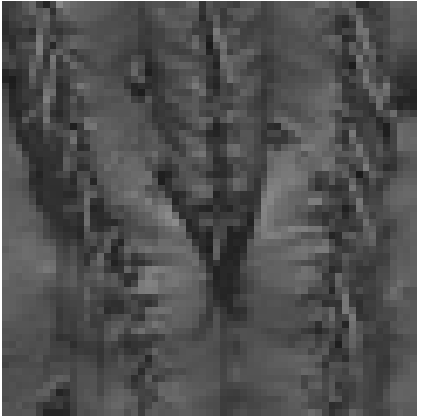} &
		\includegraphics[height=0.05\linewidth]{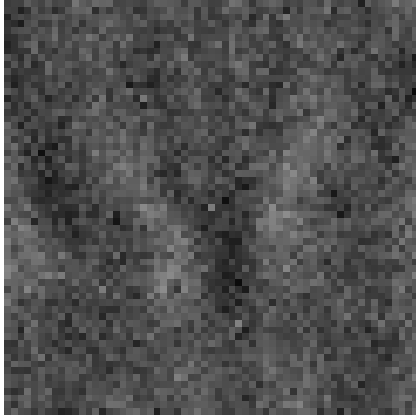} &
		\includegraphics[height=0.05\linewidth]{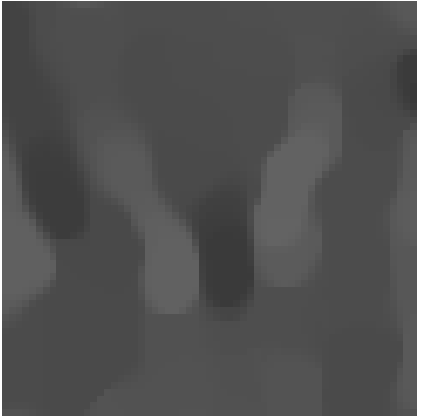} &
		\includegraphics[height=0.05\linewidth]{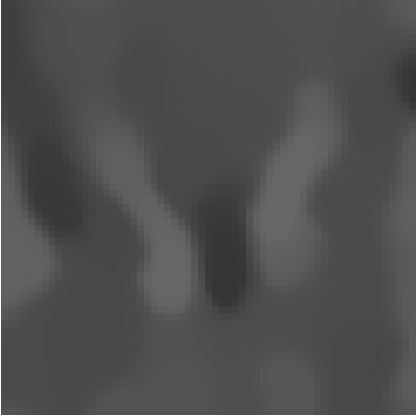} & 
		\includegraphics[height=0.05\linewidth]{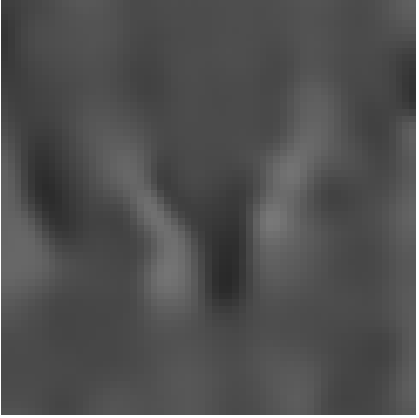} &
		\includegraphics[height=0.05\linewidth]{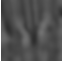} &	
		\includegraphics[height=0.05\linewidth]{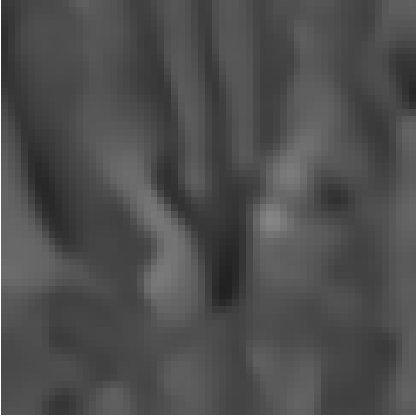} &
		\includegraphics[height=0.05\linewidth]{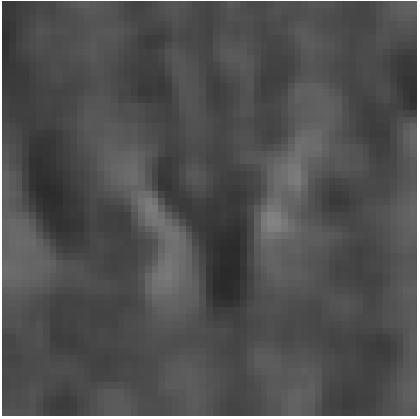}	\\
		\includegraphics[height=0.07\linewidth]{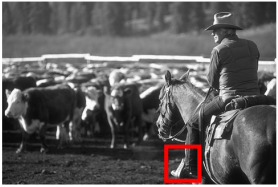} &
		\includegraphics[height=0.07\linewidth]{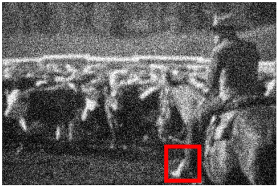} &
		\includegraphics[height=0.07\linewidth]{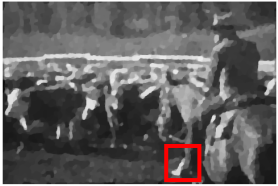} &
		\includegraphics[height=0.07\linewidth]{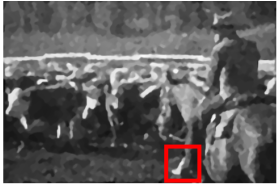} & 
		\includegraphics[height=0.07\linewidth]{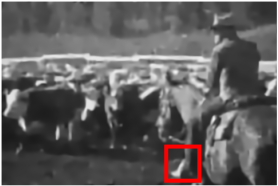} &
		\includegraphics[height=0.07\linewidth]{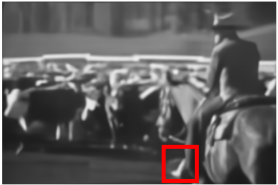} &	
		\includegraphics[height=0.07\linewidth]{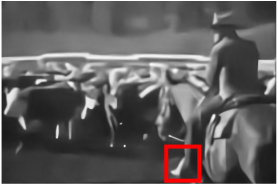} &
		\includegraphics[height=0.07\linewidth]{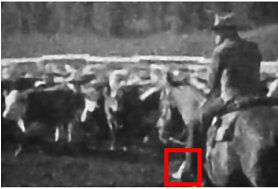} \\
		\includegraphics[height=0.05\linewidth]{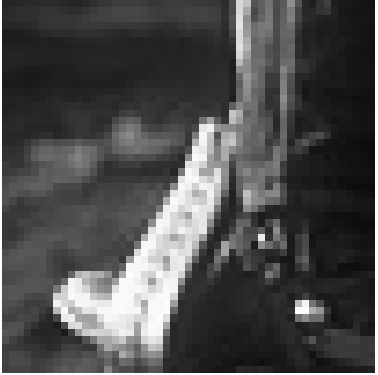} &
		\includegraphics[height=0.05\linewidth]{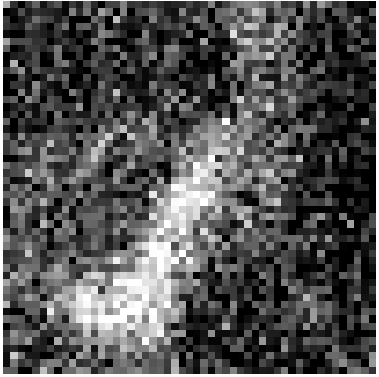} &
		\includegraphics[height=0.05\linewidth]{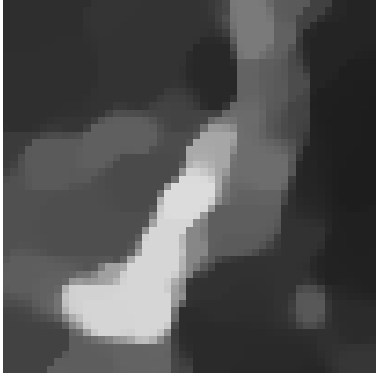} &
		\includegraphics[height=0.05\linewidth]{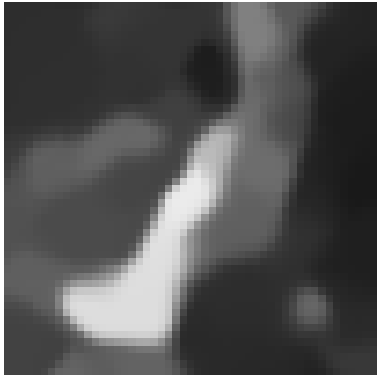} & 
		\includegraphics[height=0.05\linewidth]{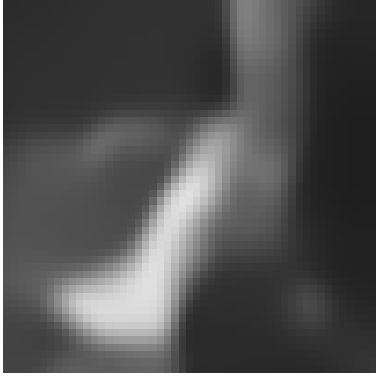} &
		\includegraphics[height=0.05\linewidth]{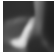} &	
		\includegraphics[height=0.05\linewidth]{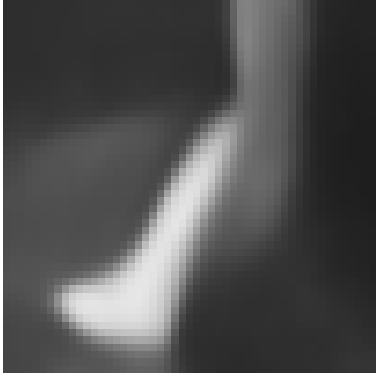} &
		\includegraphics[height=0.05\linewidth]{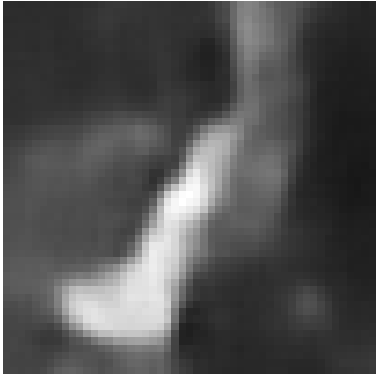} \\	
	\end{tabular}
	\caption{Visual comparisons on BSD68 dataset for different methods. The first row corresponds to the BSD68 data with a uniform $3\times3$ blur and a Gaussian noise with $\alpha=15$, the second row is the zoomed regions of the red rectangles in the first row; the third row is with a uniform $5\times5$ blur and a Gaussian noise with $\alpha=15$, the fourth row is the zoomed regions of the red rectangles in the third row; the fifth row is with a uniform $5\times5$ blur and a Gaussian noise with $\alpha=50$, the sixth row is the zoomed regions of the red rectangles in the fifth row. For each instance, the images from the first to the seventh column respectively correspond to the original image $\bar{\textrm{x}}$, the degraded one $\textrm{z}$, the restored ones by TV, NLTV, EPLL, MWCNN, IRCNN and the proposed full DeepPDNet ($Q=10, K=20$).} \label{fig:bsdinstance}
\end{figure*}

\subsection{The performance on BSD68 dataset} \label{sec:bsd68}
\noindent In this section, we evaluate the performance of the proposed DeepPDNet\footnote{Full DeepPDNet with filters ``f5s2n20''+``f7s3n20''+``f14s7n20''+``f28s28n10''} on the BSD68 dataset containing 68 natural images from Berkeley segmentation dataset with 500 images~\cite{Roth2009Fields}. We follow \cite{Chen2017Trainable} to use 400 images from BSD dataset of size 180$\times$180 to train the network, and the 68 images are chosen from the BSD data for evaluation without the overlaps with the training set. In this work, we focus on the gray version of the BSD68 dataset for restoration. In the experiments, we take into account two different blur sizes ($3 \times 3$ and $5 \times 5$) and three Gaussian noise levels ($\sigma=15, 25, 50$). Few instances of the test images are displayed in Fig.~\ref{fig:bsdinstance}.

Considering that test images in BSD68 dataset are of size 321$\times$481, it is particularly difficult to make $\mathbf{z}$ be the whole image due to the huge memory demanding. We adopt a patch-based strategy that we train an affordable network by full learning from a collection of the patches of size $Q\times Q$ randomly extracted from training images with a number of about 200K, and then the learned network is traversed over a test image to obtain the deblurred image.

The comparison results with other methods are shown in Tab.~\ref{tab:bsdfinalcomp}. From the table, we can see that: i) due to a more complex dataset, the performance of different methods are worse than the ones on the MNIST dataset; ii) We first train a network of the same architecture with the one used on the MNIST dataset, i.e.~$Q=28$, it is observed that the results are poor, especially for SSIM. We conjecture that the 6-layer network is not sufficient to express the complexity contents of the BSD dataset, therefore, we adopt a deeper network with 20 layer when $Q=10$ which is easily affordable on the experimental platform, leading to a Full DeepPDNet with filters ``f5s2n20''+``f7s3n20''+``f10s10n20''. We clearly observe that with deeper architecture, the performance are able to be further improved, especially for SSIM. Our proposed method obtains comparable results to MWCNN and IRCNN, which is based on a powerful deep convolutional network, allowing to capture the statistical property of the image; iii)~the gap between different methods become less as the noise level becomes large. The instances of the deblurred images are shown in Fig.~\ref{fig:bsdinstance}. The learned filters are shown in Fig.~\ref{fig:bsdlearnedpattern}, we can see that the learned filters are meaningful to capture the local property.

\subsection{Extension to Single Image Super-Resolution task}
In this section, we investigate the adaptability of the proposed network on the single image super-resolution task. The goal of image super-resolution task is to generate a corresponding high-resolution (HR) image from a low-resolution (LR) image. Recently, the CNN-based deep networks  have obtained promising performance by supervised learning. In the proposed method, in order to facilitate the modelization of $A$ we restrict it to downsampling while state-of-the-art methods benefit from an input obtained from a bicubic interpolation.
According to the results on BSD68 dataset in Section~\ref{sec:bsd68}, the proposed network is a Full DeepPDNet with filters ``f5s2n20''+``f7s3n20''+``f10s10n20'' and $K=20$. The learning stage is performed on the BSD200 dataset and is evaluated in the BSD100 and SET14 datasets used in~\cite{KimLeeLee_vdsr_cvpr16} with downscaling factors 2, 3 and 4.

\begin{figure*}[tbp]
	\centering
	\begin{tabular}{llllll}
		{\large \;\;\;\;\;\;\;\;$\bar{\textrm{x}}$} & \;\;\;\;\;\;VDSR & \;\;\;\;MWCNN & \;\;\;\;\;DnCNN &\;\;\;\; SRCNN & Full DeepPDNet \\
		\includegraphics[height=0.08\linewidth]{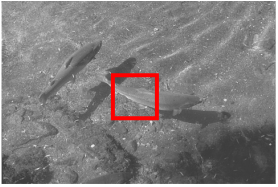} &
		\includegraphics[height=0.08\linewidth]{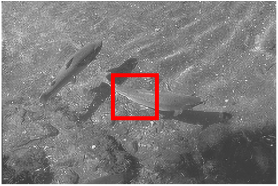} &
		\includegraphics[height=0.08\linewidth]{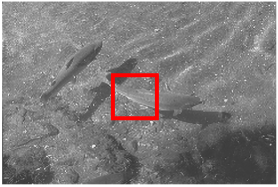} &
		\includegraphics[height=0.08\linewidth]{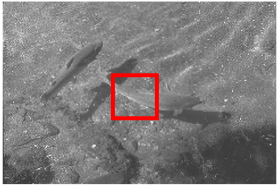} & 
		\includegraphics[height=0.08\linewidth]{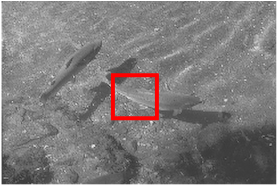} &
		\includegraphics[height=0.08\linewidth]{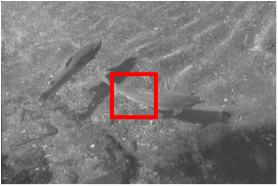} \\	
		\includegraphics[height=0.05\linewidth]{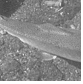} &
		\includegraphics[height=0.05\linewidth]{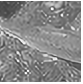} &
		\includegraphics[height=0.05\linewidth]{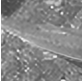} &
		\includegraphics[height=0.05\linewidth]{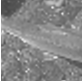} & 
		\includegraphics[height=0.05\linewidth]{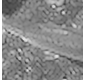} &
		\includegraphics[height=0.05\linewidth]{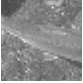} \\		
		\includegraphics[height=0.08\linewidth]{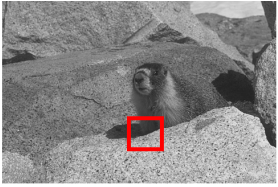} &
		\includegraphics[height=0.08\linewidth]{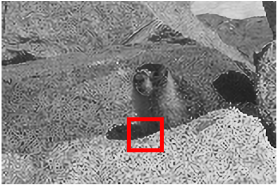} &
		\includegraphics[height=0.08\linewidth]{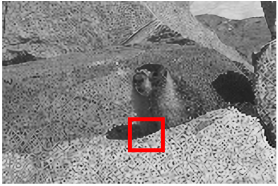} &
		\includegraphics[height=0.08\linewidth]{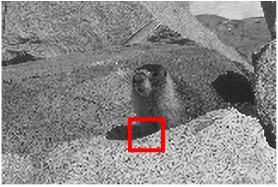} & 
		\includegraphics[height=0.08\linewidth]{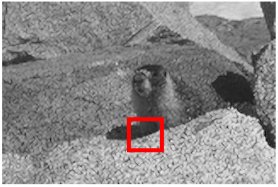} &
		\includegraphics[height=0.08\linewidth]{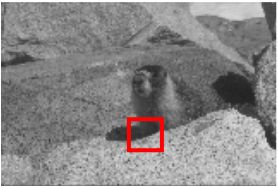} \\	
		\includegraphics[height=0.05\linewidth]{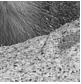} &
		\includegraphics[height=0.05\linewidth]{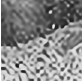} &
		\includegraphics[height=0.05\linewidth]{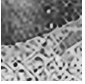} &
		\includegraphics[height=0.05\linewidth]{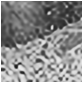} & 
		\includegraphics[height=0.05\linewidth]{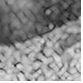} &
		\includegraphics[height=0.05\linewidth]{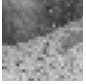} \\		
		\includegraphics[height=0.08\linewidth]{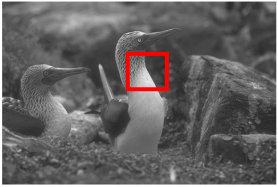} &
		\includegraphics[height=0.08\linewidth]{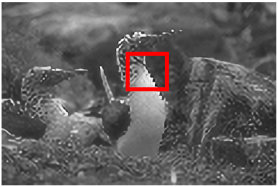} &
		\includegraphics[height=0.08\linewidth]{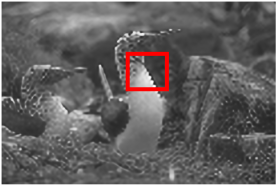} &
		\includegraphics[height=0.08\linewidth]{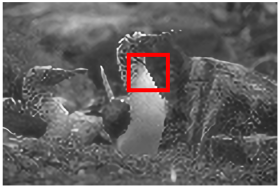} & 
		\includegraphics[height=0.08\linewidth]{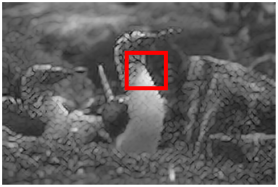} &
		\includegraphics[height=0.08\linewidth]{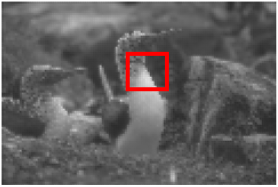} \\
		\includegraphics[height=0.05\linewidth]{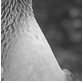} &
		\includegraphics[height=0.05\linewidth]{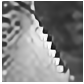} &
		\includegraphics[height=0.05\linewidth]{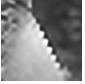} &
		\includegraphics[height=0.05\linewidth]{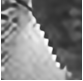} & 
		\includegraphics[height=0.05\linewidth]{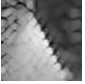} &
		\includegraphics[height=0.05\linewidth]{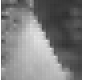} \\		
	\end{tabular}
	\caption{Visual comparisons on BSD100 dataset for different methods. The first row is the instances with 2 upscaling factor and the second row is the zoomed regions of the red rectangles in the first row; the third row is the instances with 3 upscaling factor and the forth row is the zoomed regions of the red rectangles in the third row, the fifth row is the instances with 4 upscaling factor and the sixth row is the zoomed regions of the red rectangles in the fifth row.} \label{fig:bsdssisrinstance}	
\end{figure*}

The performance comparisons are shown in Tab.~\ref{tab:bsdsisrcomp}. We compare the proposed algorithm with other CNN-based methods, for instance, SRCNN~\cite{DongChen_srcnn_eccv14}, VDSR~\cite{KimLeeLee_vdsr_cvpr16}, DnCNN~\cite{Zhang2017}, and MWCNN~\cite{LieZhang-wmcnn_cvpr18}\footnote{The test codes and released models are also downloaded from the authors' websites.}. SRCNN~\cite{DongChen_srcnn_eccv14} directly learns an end-to-end mapping between the low/high resolution by using a deep CNN. Similar to VGG-net, VDSR~\cite{KimLeeLee_vdsr_cvpr16} learns a very deep CNNs of 20 layers for accurate single-image super-resolution. DnCNN~\cite{Zhang2017} adopts residual learning to predict the difference between the high-resolution ground truth and the bicubic upsampling of a low-resolution image, and MWCNN~\cite{LieZhang-wmcnn_cvpr18} integrates the wavelet transform to the U-Net architecture to obtain mid-level wavelet-CNN framework. We also propose to retrain MWCNN and DnCNN models, for which training codes are available. The training stage takes as an input $\textrm{z}_s$, obtained from a direct down-sampling, and the first layer of the network involves a direct upsampling of $\textrm{z}_s$. Similar procedure is considered for the test stage. From Table~\ref{tab:bsdsisrcomp}, it can be seen that: i) our method outperforms the released models of state-of-the-art methods; ii) for both re-trained MWCNN and DnCNN models, the performance are improved compared to their counterpart released models; iii) our method always outperforms the re-trained MWCNN for both BSD100 and SET14 datasets, iv) the performance of Full DeepPDNet and the re-trained DnCNN are comparable. Fig.~\ref{fig:bsdssisrinstance} also visualizes the image instances by different methods.

\begin{table}[t]
	\centering
	\resizebox{\linewidth}{!}{
		\begin{tabular}{|c|c|c|c|c|}
			\hline
			\multirow{2}{*}{Dataset} & \multirow{2}{*}{Method} & \multicolumn{3}{c|}{Upscaling Factor} \\
			\cline{3-5}
			& & 2 & 3 & 4 \\
			\cline{3-5}			
			& & PSNR/SSIM & PSNR/SSIM & PSNR/SSIM \\
			\hline
			\multirow{7}{*}{BSD100} & VDSR~\cite{KimLeeLee_vdsr_cvpr16} & 24.12/0.7242 & 21.09/0.5656 & 19.66/0.4896 \\
			& MWCNN~\cite{LieZhang-wmcnn_cvpr18} & 23.66/0.6733 & 21.91/0.6078 & 21.38/0.5541 \\
			& DnCNN~\cite{Zhang2017} & 27.15/0.8089 & 22.12/0.6131 & 21.53/0.5803 \\
			& SRCNN~\cite{DongChen_srcnn_eccv14} & 26.63/0.7908 & 23.88/0.6469 & 22.42/0.5629 \\
			& Re-trained MWCNN~\cite{LieZhang-wmcnn_cvpr18} &{28.59}/0.8306 & 21.70/0.5987 & 22.32/0.5958\\
			& Re-trained DnCNN~\cite{Zhang2017} &  27.25/{0.8439} & {25.54}/{0.7315} & {23.99}/{0.6591} \\
			& Full DeepPDNet & {28.88/0.8414} & {25.51/0.7155} & {23.93/0.6193} \\
			\hline
			\multirow{7}{*}{SET14} & VDSR~\cite{KimLeeLee_vdsr_cvpr16} & 24.40/0.7528 & 20.71/0.5885 & 18.97/0.5042\\
			& MWCNN~\cite{LieZhang-wmcnn_cvpr18} & 22.93/0.6745 & 20.71/0.5801 & 19.52/0.5168 \\
			& DnCNN~\cite{Zhang2017} & 24.95/0.7653 & 22.27/0.6601 & 19.99/0.5518 \\	
			& SRCNN~\cite{DongChen_srcnn_eccv14} & 25.65/0.7819 & 24.07/0.6906 & 21.12/0.5417 \\
			& Re-trained MWCNN~\cite{LieZhang-wmcnn_cvpr18} &{28.78}/0.8479 & 21.53/0.6201 & 21.64/0.6013\\
			& Re-trained DnCNN~\cite{Zhang2017} &  {29.07}/{0.8594} & {25.03}/{0.7568} & {24.51}/{0.6942} \\			
			& Full DeepPDNet & 28.58/{0.8618} & {25.45/0.7453} & {23.17/0.6411} \\		 
			\hline
		\end{tabular}
	}
	\caption{Comparison results of different methods on the BSD100 and SET14 dataset for single image super-resolution task for different upscaling factors.} \label{tab:bsdsisrcomp}
\end{table}

\section{Conclusion} \label{sec:conclusion}
This work aims to design a flexible network using prior knowledge on inverse problems both in terms of penalized co-log-likelihood design and optimization schemes. Our contribution is first to unfold the PDHG iterations and to establish a connection with standard neural networks with $K$ layers. From this preliminary network, we design DeepPDNet that allows us to learn both the algorithmic step-size and the analysis linear operator  (including implicitly the regularization parameter) involved in each layer. A full and a partial DeepPDNet are provided, one considering the learning of all the parameters without constraints leading to better numerical performance, and a second one allowing to provide a framework more adapted to theoretical stability analysis. Dense (i.e. global) or block-sparse (i.e. local) sparse analysis prior operators are considered. The backpropagation procedure is detailed. We employ the proposed network for two types of degradation: image deblur and single image super-resolution task, the experimental results illustrate that for images with small complexity such as MNIST, a network with few layers allows us to outperforms state-of-the-art methods while for a more complex dataset, the proposed method outperforms standard unsupervised approaches such as TV, NLTV or EPLL and achieves comparable results with state-of-the-art CNN based methods. Moreover, we observed that the more complex is the dataset, the deeper needs to be the network. 

Such a procedure can be extended to other type of degradation model considering for instance weighted data-term or by integrating only partial knowledge of degradation process in order to perform  blind deconvolution following ideas provided in~\cite{RenZuoZhangPAMI2019}.

\section*{Appendix A: Computation of derivatives} \label{app:gradientcomp}

\begin{enumerate}
	\item 
	$\frac{\partial \textrm{u}^{[k+1]}_s}{\partial \textrm{c}^{[k]}_s}$ is the derivative of output $\textrm{u}^{[k+1]}_s$ at layer $k+1$ w.r.t. $\textrm{c}^{[k]}_s = (\textrm{c}^{[k]}_{s,1},\textrm{c}^{[k]}_{s,2})$ according to \eqref{eq:forward2}. Since identity acts on $\textrm{c}^{[k]}_{s,1}$ and proximity operator act on $\textrm{c}^{[k]}_{s,2}$, this yields to:
	\begin{equation}
	\frac{\partial \textrm{u}^{[k+1]}_s}{\partial \textrm{c}^{[k]}_s}  = (\textrm{u}_1^\top,\textrm{u}_2^\top)^\top \in \mathbb{R}^{N} \times \mathbb{R}^{P}
	\end{equation}
	\noindent where $\textrm{u}_1 = (1,\ldots,1)$  and where $\textrm{u}_{2,p}$ when $g= \Vert \cdot \Vert_1$ is:
	\begin{equation}
	\textrm{u}_{2,p} = \begin{cases}
	0 &  \mbox{if}\;\vert \textrm{c}^{[k]}_{s,2,p}\vert >1 \\
	1 &  \mbox{if}\;\vert \textrm{c}^{[k]}_{s,2,p}\vert <1 \\
	[0,1] &\mbox{if}\; \textrm{c}^{[k]}_{s,2,p} =\pm1.
	\end{cases} \label{eq:subgradproximall1}
	\end{equation}
	\item For the middle layers except the first and last layer, the derivative of the elementary of the network parameters ($D^{[k]}$, $b^{[k]}$) w.r.t. $\tau^{[k]}$, $\sigma^{[k]}$, $L^{[k]}$ (i.e.~$\frac{\partial b^{[k]}_{[\ell]}}{\partial \tau^{[k]}_{[\ell]}}$, $\frac{\partial D^{[k]}_{[\ell]}}{\partial \tau^{[k]}_{[\ell]}}$, $\frac{\partial b^{[k]}_{[\ell]}}{\partial \sigma^{[k]}_{[\ell]}}$, $\frac{\partial D^{[k]}_{[\ell]}}{\partial \sigma^{[k]}_{[\ell]}}$, $\frac{\partial b^{[k]}_{[\ell]}}{\partial L^{[k]}_{[\ell]}}$ and $\frac{\partial D^{[k]}_{[\ell]}}{\partial L^{[k]}_{[\ell]}}$) can be easily obtained from Eq.~\eqref{equa:paramdef}:
	\begin{equation} \label{eq:derivativelist}
	\begin{cases}
	\frac{\partial b^{[k]}_{[\ell]}}{\partial \tau^{[k]}_{[\ell]}} & = \begin{pmatrix}
	A^*\textrm{z}\\
	2 \sigma^{[k]}_{[\ell]} L^{[k]}_{[\ell]} A^*\textrm{z}\\
	\end{pmatrix} \\
	\frac{\partial D^{[k]}_{[\ell]}}{\partial \tau^{[k]}_{[\ell]}} & = \begin{pmatrix}
	- A^*A & - L^{[k]*}_{[\ell]} \\
	-2 \sigma^{[k]}_{[\ell]} L^{[k]}_{[\ell]} A^*A & - 2 \sigma L^{[k]}_{[\ell]} L^{[k]*}_{[\ell]}
	\end{pmatrix} \\
	\frac{\partial b^{[k]}_{[\ell]}}{\partial \sigma^{[k]}_{[\ell]}} & =  \begin{pmatrix}
	0_{N \times 1} \\
	2\tau^{[k]}_{[\ell]}  L^{[k]}_{[\ell]} A^*\textrm{z}\\
	\end{pmatrix} \\
	\frac{\partial D^{[k]}_{[\ell]}}{\partial \sigma^{[k]}_{[\ell]}} & = \begin{pmatrix}
	0_{N \times N}  & 0_{N \times P} \\
	L^{[k]}_{[\ell]}(\textrm{Id} - 2\tau^{[k]}_{[\ell]} A^*A) & - 2\tau^{[k]}_{[\ell]} L^{[k]}_{[\ell]} L^{[k]*}_{[\ell]} \\
	\end{pmatrix} \\
	\frac{\partial b^{[k]}_{[\ell]}}{\partial L^{[k]}_{[\ell]}} & = \begin{pmatrix}
	0_{N\times 1} \\
	2\tau^{[k]}_{[\ell]} \sigma^{[k]}_{[\ell]} A^*\textrm{z}\\
	\end{pmatrix} \\
	\frac{\partial D^{[k]}_{[\ell]}}{\partial L^{[k]}_{[\ell]}} & = \begin{pmatrix}
	0_{N \times N} & - \tau^{[k]}_{[\ell]} \\
	\sigma^{[k]}_{[\ell]} (\textrm{Id} - 2\tau^{[k]}_{[\ell]} A^*A) & -4 \tau^{[k]}_{[\ell]} \sigma^{[k]}_{[\ell]} L^* \\
	\end{pmatrix} \\
	\end{cases}
	\end{equation}
	
	\noindent  and then the respective gradients are accumulated from all the elementary of the $D^{[k]}$, $b^{[k]}$. Similarly, the corresponding derivatives in the first and last layer are similarly calculated according to Eq.~\eqref{eq:newstructure} and Eq.~\eqref{eq:derivativelist}.
	
	\item $\frac{\partial u^{[k]}_{s+1}}{\partial \sigma^{[k]}_{[\ell]}}$ is the gradient of conjugate of proximity operator $\ell_1$-norm w.r.t. $\sigma^{[k]}$ at iteration $\ell$:
	\begin{equation}
	\frac{\partial u^{[k+1]}_{s}}{\partial \sigma^{[k]}_{[\ell]}} = 0.
	\end{equation}
\end{enumerate}

\section*{Acknowledgement}
This work is supported by the National Natural Science Foundation of China (61806180, U1804152), Key Research Projects of Henan Higher Education Institutions in China (19A520037), Science and Technology Innovation Project of Zhengzhou (2019CXZX0037), and also by the ANR (Agence Nationale de la Recherche) of France (ANR-19-CE48-0009 Multisc'In).


\end{document}